\documentclass{article}

\usepackage[preprint]{neurips_2026}

\usepackage[utf8]{inputenc}
\usepackage[T1]{fontenc}
\usepackage{microtype}
\usepackage{booktabs}
\usepackage{float}
\usepackage{enumitem}
\usepackage{algorithm}
\usepackage{algorithmic}
\usepackage{amsmath}
\usepackage{amssymb}
\usepackage{amsthm}
\usepackage{graphicx}
\usepackage{xcolor}
\usepackage{tikz}
\usetikzlibrary{positioning,shapes.geometric}
\usepackage{hyperref}

\newtheorem{proposition}{Proposition}
\newtheorem{openproblem}{Open Problem}

\title{TRACE: An Operational Reasoning Schema for Auditable Agentic Commitments}

\author{%
  Edward Y.\ Chang\\
  Stanford University\\
  \texttt{echang@cs.stanford.edu}
  \And
  Emily J.\ Chang\\
  Quadrivium AI
}

\begin{document}

\maketitle

\begin{abstract}
This paper defines and proposes TRACE (Typed Reasoning And Commitment Evidence): a typed, versioned schema for recording reasoning traces,
one reference procedure for writing records against that schema, and one operating discipline, no durable state change without a record. The paper first states goals and non-goals, then argues in three layers that reasoning is not in the language model: the autoregressive mechanism natively computes association; chain-of-thought and reinforcement learning inherit the mechanism's limits; and the formal constructs of reasoning theory, from Socratic procedure to Pearl's ladder, are absent as machinery. The TRACE schema answers the absence with fields and tests: the TraceRecord and its causal specialization, an eight-stage reference writer, a gate-first measurement regime, the TRACE-Bench protocol, and the consumers, memory admission, plan gating, temporal regret, and verdict reuse, whose improved and more auditable decisions are the measure by which the record is judged. A record-consumer contract states the guarantees a record makes to a consumer and the obligations a consumer accepts in return, so the schema is an operational interface rather than a passive document. Two worked
examples run in the main text: a music-lessons argument traced from raw sentence to typed verdict, which shows the writer separating association, intervention, and prescription; and a flood search-and-rescue vignette in which a predictive world model reports a confident plan success that its own support and out-of-distribution scores contradict, so the record defers the commitment, requests a bounded observation, revises the claim append-only, and clears a different plan branch. The vignette is illustrative, not an empirical result; a closed-loop evaluation is specified but left to future work, so the paper's contribution is the schema and its consumer contract, not a
performance claim. Appendices carry the full schema, the reference writer's algorithms and cost model, the contested anatomy, worked illustrations at clinical and policy stakes, the benchmark protocol, the convergence-validity metrics and imported propositions, and usage scenarios for memory admission and plan gating.
\end{abstract}

\section{The Paper's Aim: TRACE, Its Goals, and Its Non-Goals}
\label{sec:ocr-problem}

This paper defines and proposes TRACE (Typed Reasoning And Commitment Evidence): a typed, versioned schema for recording reasoning traces,
one reference procedure for writing records against that schema, and one
operating discipline that makes the records consequential. The artifact is
the TRACE record; everything else in the paper either writes it, measures
it, or consumes it. So that the reader's expectations are set before any
machinery arrives, the paper's goals and non-goals are stated here, first.

\paragraph{Goals.} The paper pursues three.
\begin{enumerate}
    \item \emph{Record.} Define one unified schema, concretely a versioned
    JSON type definition, that captures the typed structure of a reasoning
    episode: what was claimed, what kind of reasoning the claim required,
    what evidence and warrants were offered, what was tested, what failed,
    what is missing, and what verdict followed.
    \item \emph{Protocol.} Make the same schema the interchange format
    between an agentic system's services: writers emit records, consumers act on
    them, and no service passes reasoning state to another in any other
    form. The schema is simultaneously a log format and an exchange
    protocol.
    \item \emph{Improvement, through the callers.} Improve reasoning
    quality where it can actually be improved: in the systems that call
    TRACE and consume its records. SocraSynth \cite{chang2024socrasynth} and EVINCE \cite{chang2024evince} elicit against
    recorded claims; RCA \cite{chang2026sycophancy} localizes failures to recorded stages; ERM \cite{chang2026erm} prices
    recorded errors; RLER \cite{chang2026pathagi2} revises policies from recorded verdicts; Trivium \cite{chang2026trivium}
    computes temporal regret over recorded steps; Mnemosyne \cite{chang2027mnemosyne} gates memory
    and MACI \cite{chang2024pathagi1} gates plans on recorded outcomes. The record is what makes
    each of these operations possible, and consumer value is how the record
    is judged.
\end{enumerate}

\paragraph{Non-goals.} Three disclaimers bound the proposal.
\begin{enumerate}
    \item TRACE does not itself optimize or improve reasoning. It is
    infrastructure in service of its callers: it records and adjudicates so
    that the systems above it can improve reasoning, the way a write-ahead
    log recovers no transaction by itself.
    \item The eight-stage reference writer is not claimed to be optimal;
    it is one conforming implementation, and every stage is replaceable
    without moving the record format.
    \item TRACE is not a universal reasoning-quality metric. Its scores
    measure compliance with the declared, versioned TRACE policy; no
    objective scalar of reasoning quality exists, and nothing essential
    in the design depends on one.
\end{enumerate}

\paragraph{Two precedents.} Systems engineering has met this situation
twice before. A write-ahead log does not make a database faster or smarter,
and a flight recorder does not fly the aircraft. Yet recovery, replication,
and audit are impossible without the first, post-incident analysis is
impossible without the second, and in both cases the value of the log is
proven entirely by what its consumers can do. The TRACE record plays the
same role for reasoning in an agentic system. It does not make the
underlying model more correct. It makes every durable consequence of the
model's reasoning attributable, contestable, and improvable.

\paragraph{The TRACE discipline.} The schema comes with one operating
rule, which the companion implementation enforces: no durable state change
without a TRACE record. A memory is not committed, a plan is not cleared, and an
obligation is not updated unless a TRACE record has been written against
the schema and carries a verdict. Section~\ref{sec:ocr-flood-vignette}
makes this closure concrete in a flood search-and-rescue plan: a high
prediction is held when its support fails, a drone observation repairs the
record, and only the dependent route branch is revised. The companion
volume's implementation chapter closes this loop; Appendices~\ref{app:mnemosyne}
and~\ref{app:maci} sketch the two admission consumers.

The paper's design claim is therefore:

\begin{quote}
The unit of contribution is the record, not any stage algorithm. Reasoning
evaluation should be decomposed into typed, auditable responsibilities that
read and write one versioned TRACE record; a claim must be typed before
it is tested, and tested at the standard its type demands; and the record is
judged not by a quality scalar but by what its consumers,
admission control, plan gating, and temporal learning, can do with it.
\end{quote}

Two consequences of this claim are worth stating before the machinery
arrives. First, the paper's foundational claim is not that TRACE measures
reasoning quality; it is that TRACE records the typed structure of reasoning
sufficiently for admission, audit, regret, repair, and downstream learning.
Throughout this paper and its companion volume, TRACE names this whole contract: the TRACE schema,
the TRACE writer contract, and the TRACE discipline, so that a TRACE, in the
ordinary sense of the word, is exactly what the system leaves behind. TRACE
still performs evaluation-like functions, but its verdicts are fields of the
record, and the stages presented below are one implementation of the writer
contract, every one of them replaceable without moving the record format.
Second, the scores defined later measure compliance with the declared,
versioned TRACE policy, not objective reasoning quality.
Section~\ref{sec:ocr-falsify} states both points precisely, together with
what would falsify the design.

\paragraph{Contributions.} The paper makes five contributions. First, a
\emph{schema}: the TraceRecord and its causal specialization the
TraceCausalRecord (Section~\ref{sec:ocr-record}), versioned like an
interface, carrying evaluation state, including rung, identification status,
bounds, missing evidence, and regret, rather than argument structure alone.
Second, a \emph{type system} for the schema's fields: four layers of typed
evaluation that route claims across formal disciplines (identification
theory, proof checking, decision theory, law), where prior argument-mining
pipelines and scheme-based systems such as Carneades classify and evaluate
within the argumentation layer; an explicit subsumption rule governs which
layer adjudicates when they overlap. Third, a \emph{reference writer}
(Algorithms~\ref{alg:ocr-trace} and~\ref{alg:ocr-trace2}), one conforming
implementation of the writer contract, presented for auditability and cost
discipline rather than claimed as optimal. Fourth, \emph{two
complementary worked illustrations}: one ordinary sentence followed through
every writer stage, and one flood search-and-rescue plan followed from
predictive evidence through defer, evidence-seeking repair, append-only
revision, and plan-gate consumption. Fifth, a \emph{benchmark and ablation
plan}, TRACE-Bench, which judges the
schema on expressiveness, the writer on stage-level and end-to-end quality,
and the record on consumer value, and whose baselines include the
single-pass writer that serves as the null hypothesis for the staged
decomposition.

\paragraph{Organization.} The paper proceeds as aim, critique, schema,
hierarchy, stages, measurement, consumers, illustrations, debate,
guarantees, and falsification. Section~\ref{sec:ocr-why-reasoning} states
the layered critique. Section~\ref{sec:ocr-record} fixes the terminology,
the schema and its key fields, and the writer API
(Section~\ref{sec:ocr-trace-method}). Section~\ref{sec:ocr-hierarchy}
builds the reasoning hierarchy that gives the typing fields their
semantics: the CRIT roof (Section~\ref{sec:ocr-reasoning-vs-debate}), the
four evaluation layers (Section~\ref{sec:ocr-four-layers}), the reasoning
families with the Toulmin, Walton, and Dung toolbox and the
enthymeme-reconstruction algorithm
(Section~\ref{sec:ocr-reasoning-families}), and Pearl's ladder as the
causal plug-in (Section~\ref{sec:ocr-pearl-map}).
Section~\ref{sec:ocr-stages} depicts the writer's stages: the formulation
gate (Section~\ref{sec:ocr-detection}), one claim traced from sentence
to verdict (Section~\ref{sec:ocr-endtoend}), and the stopping rules
(Section~\ref{sec:ocr-stopping}). Section~\ref{sec:ocr-measurement}
defines the TRACE policy's quality metrics and hard gates; the benchmark
that operationalizes them, TRACE-Bench, is Appendix~\ref{app:bench}, and
the convergence theory and validity metrics of the debate stage are
Appendix~\ref{app:convergence}. Section~\ref{sec:ocr-falsify} states what the paper
claims, including the consumer claim that TRACE is a better admission
explainer rather than a better gate. Section~\ref{sec:ocr-flood-vignette}
then gives the operational consumer illustration before the paper states what
it does not claim, what would falsify it, and its limits. Worked
illustrations at clinical and policy stakes are in
Appendix~\ref{app:illustrations}.

\section{Why Reasoning Is Not in the Language Model}
\label{sec:ocr-why-reasoning}

Reasoning is the capability on which the current era of AI turns. The decisive
advances of the last five years were not simply advances in knowledge storage.
They were advances in \emph{elicited computation}: training models to emit
intermediate steps or scratchpads~\cite{nye2021scratchpads}, prompting them to
produce chains of thought~\cite{wei2022chain,kojima2022zeroshot}, sampling and
voting over diverse reasoning paths~\cite{wang2023selfconsistency},
decomposing problems from least to most complex~\cite{zhou2023least}, searching
over branching thoughts~\cite{yao2023tree}, and interleaving reasoning with
action~\cite{yao2023react}. The industry then made reasoning the explicit
scaling axis: allocating inference-time compute to deliberation can outperform
scaling parameters~\cite{snell2025scaling}, and reinforcement learning with
verifiable rewards produces models that plan, self-check, and revise before
answering~\cite{openai2024o1,guo2025deepseekr1}. Surveys of this trajectory
now describe a shift from prompting techniques to ``large reasoning
models''~\cite{chang2023crit,chen2025reasoningera,huang2023towards,qiao2022reasoning}.

Before the schema can be motivated, the critique it answers must be stated
in an organized way, because the problems are of different depths and are
too often listed as one undifferentiated complaint. The critique here has
three layers, each deeper than the last. Layer one examines the mechanism:
what an autoregressive model natively computes. Layer two examines the two
current patches, chain-of-thought elicitation and reinforcement learning,
and shows that both inherit the mechanism's limits. Layer three brings in
the formal constructs of reasoning theory and asks, construct by construct,
where each one lives in a language model; the answer, nowhere, converts the
critique into the specification that the TRACE schema answers.

\subsection{Layer One: The Mechanism Is Association}
\label{sec:ocr-layer-mechanism}

Elicitation, however, changes what the model emits, not what the model
computes. Maximum-likelihood training makes association the native
operation: an autoregressive model is optimized to continue text as its
corpus would, and correlation, not causation, is what corpus statistics
carry. When a model's output appears causal, it is because causally
generated text was absorbed associatively, the corpus contains the
sentences that causal processes in the world produced, not because the
autoregressive mechanism can distinguish a causal relation from an
association. Models accordingly deploy causal vocabulary without satisfying
causal semantics~\cite{kiciman2024causal,zecevic2023parrots}: the words
\emph{because}, \emph{causes}, and \emph{therefore} are high-probability
tokens, not inference steps.

The claim should be stated at its defensible strength. Mechanistic
interpretability has found nontrivial intermediate structure inside trained
transformers, including world-model-like state representations and evidence
of forward planning~\cite{li2023othello,lindsey2025biology}, so the layer's
argument is not that no internal structure exists. The argument is narrower
and sufficient: the training objective neither requires nor certifies causal
semantics, and whatever structure emerges is not exposed as checkable
machinery, no typed claim, no identification status, no auditable warrant.
A schema is needed precisely because the mechanism offers no interface at
which such properties could be verified, whatever is happening inside.

\paragraph{The maximum-likelihood trap.} The point has a distributional
statement. Autoregressive decoding often favors high-probability
continuations under
\(P_\theta(\text{continuation}\mid\text{context})\), a distribution shaped by
corpus frequency, instruction tuning, and preference optimization rather than
by evidential support. Decoding strategies vary, but decoding alone does not
distinguish evidential support from high-probability continuation unless
evidence is explicitly represented, weighted, and audited. Call the
\emph{modal basin} the region of the model's conditional distribution
containing familiar, high-probability continuations that tend to reinforce
one another during generation. Reasoning that remains inside this basin tends
to reproduce the familiar answer rather than the best-supported conclusion:
the majority view, the standard explanation, the safest hedge.

Conditioning is the steering wheel out of the basin:
\(P_\theta(\cdot\mid\text{context},\allowbreak\text{role},\allowbreak\text{contentiousness})\) is a
different distribution, and its high-probability continuations may have low
probability under the unconditioned distribution. Adversarial role assignment
therefore surfaces continuations that are coherent yet unlikely to be produced
by an unconditioned pass~\cite{chang2024socrasynth}. Whether an exchange
actually left the basin, and whether it returned for the right reasons, is a
measurable question, monitored by information-theoretic dials and claim-level
audits~\cite{chang2024evince,chang2023crit}. The trap also explains why
purely self-directed reasoning has no reliable mechanism for escaping it:
chain-of-thought, self-evaluation, and self-consistency all sample from
closely related conditionals, so their internal agreement may be a property
of the basin, not of the world.

\subsection{Layer Two: The Patches Inherit the Mechanism}
\label{sec:ocr-layer-patches}

Two patches dominate current practice: elicit a chain of thought, and train
with reinforcement on outcomes. Both help; neither changes the mechanism,
and each fails in the way the mechanism predicts.

\paragraph{Chain-of-thought.} A model can produce a plausible chain because
the words and steps are strongly associated in its training distribution,
not because the chain faithfully caused the answer. The empirical record
supports this caution. Chain-of-thought explanations can misreport the
factors that actually drive a model's answer~\cite{turpin2023language};
answers can survive truncation or corruption of the stated chain, showing
that the chain is often post hoc~\cite{lanham2023measuring}; and
reinforcement-trained reasoning models reveal exploited hints in their
visible chains only a minority of the time~\cite{chen2025reasoningmodels}.
Measured competence is also brittle: accuracy degrades under superficial
perturbations of familiar problems~\cite{mirzadeh2025gsm} and under
counterfactual variants of routine tasks~\cite{wu2024reasoning}. Reported
collapses of reasoning models at high problem
complexity~\cite{shojaee2025illusion} are themselves
disputed~\cite{lawsen2025illusion}, and the dispute is revealing: the field
lacks agreed instruments for measuring reasoning quality. Read against the
families of reasoning developed later in this paper, a chain of thought at
its best mimics one family only, abduction: it produces a plausible account
after the fact. It performs no validity check, consults no identification
condition, and surfaces no defeater, which is why fluency and correctness so
often part company.

\paragraph{Reinforcement learning.} Outcome reward addresses \emph{what}
answer was accepted while leaving unclear \emph{why} the answer was correct,
why another answer was wrong, or which hidden assumption failed \cite{chang2026erm, chang2026trivium}. The reward
is also universal: one scalar per episode, blind to the user, the stakes,
the claim type, and the evidence standard the claim required, so the same
gradient that fixes one context's error installs another context's bias. The
training signal itself is contested: reinforcement learning with verifiable
rewards may mostly re-weight reasoning paths already latent in the base
model~\cite{yue2025does}, may expand them under prolonged
training~\cite{liu2025prorl}, and can register ``gains'' under spurious
rewards~\cite{shao2025spurious}. In the worst case, the model learns to
satisfy the evaluator or the user rather than to repair its own reasoning:
models defer to users against their own
evidence~\cite{perez2023discovering,sharma2024sycophancy}, the route from
apparent learning to sycophancy \cite{chang2026sycophancy}. The layer's conclusion is blunt: if the
mechanism does not support reasoning, patches that elicit more
reasoning-like text bear no fruit, and rewarding the text's outcomes teaches
the mechanism to imitate success rather than to reason.

\subsection{Layer Three: The Formal Absence}
\label{sec:ocr-layer-absence}

The deepest layer of the critique uses the formal constructs of reasoning
theory as instruments. The object of evaluation is the reasoning relation: a
typed claim, the reasons offered for it, the hidden warrants that connect
premises to conclusion, the evidence standard that applies, the
counter-reasons that would defeat it, and the context in which the claim
will be used. Centuries of work, from argumentation theory through causal
inference, have made each part of that relation a formal construct with
checkable conditions. The hammer questions are then direct. Where, in a
language model, is elenchus? Where are the argument roles, the warrant, the
qualifier, the rebuttal? Where are the argument schemes and their critical
questions, the defeaters and the attack graph, the burden of proof, the
evidence standard that changes with the claim type? Where is the declared
causal graph? Where is \(do(x)\), the identification condition that
licenses an interventional claim, and the bounds that apply when it is not
met?

Table~\ref{tab:ocr-absence} enumerates the constructs; the answer to every
question is the same, nowhere, and each row names the section of this
paper where the TRACE schema gives the construct a field and a test.

\begin{table}[H]
\centering
\footnotesize
\begin{tabular}{p{0.36\linewidth}p{0.42\linewidth}p{0.12\linewidth}}
\toprule
\textbf{Formal construct} & \textbf{What it provides} & \textbf{Section(s)} \\
\midrule
Socratic methods (elenchus, hypothesis elimination, maieutic, dialectic) & the testing procedures that refute, eliminate, and draw out claims & Section~\ref{sec:ocr-reasoning-vs-debate} \\
\addlinespace[2pt]
Claim types and reasoning families & the standard a claim must be tested against & Section~\ref{sec:ocr-reasoning-families} \\
\addlinespace[2pt]
Argument roles (warrant, qualifier, rebuttal) & the anatomy that makes hidden premises visible & Section~\ref{sec:ocr-reasoning-families} \\
\addlinespace[2pt]
Schemes and critical questions & pre-packaged cross-examination each argument form owes & Section~\ref{sec:ocr-reasoning-families} \\
\addlinespace[2pt]
Defeaters and the attack graph & the accounting of objections and their resolution & Section~\ref{sec:ocr-reasoning-families} \\
\addlinespace[2pt]
Burden of proof and evidence standards & who must show what, and to what threshold & Sections~\ref{sec:ocr-reasoning-families}, \ref{sec:ocr-measurement} \\
\addlinespace[2pt]
Declared causal graph (DAG) & the assumptions an interventional claim rests on & Section~\ref{sec:ocr-pearl-map} \\
\addlinespace[2pt]
\(do(x)\) and identification & whether the requested effect is estimable from the declared evidence & Section~\ref{sec:ocr-pearl-map} \\
\addlinespace[2pt]
Partial-identification bounds & the honest answer when identification fails: defer with bounds & Section~\ref{sec:ocr-pearl-map} \\
\addlinespace[2pt]
Counterfactual machinery & abduction, action, prediction over a structural model & Section~\ref{sec:ocr-pearl-map} \\
\bottomrule
\end{tabular}
\caption{The formal-absence checklist. Each construct is standard in its
home discipline, absent as machinery in a language model, and supplied by
 TRACE as a field with a test.}
\label{tab:ocr-absence}
\end{table}

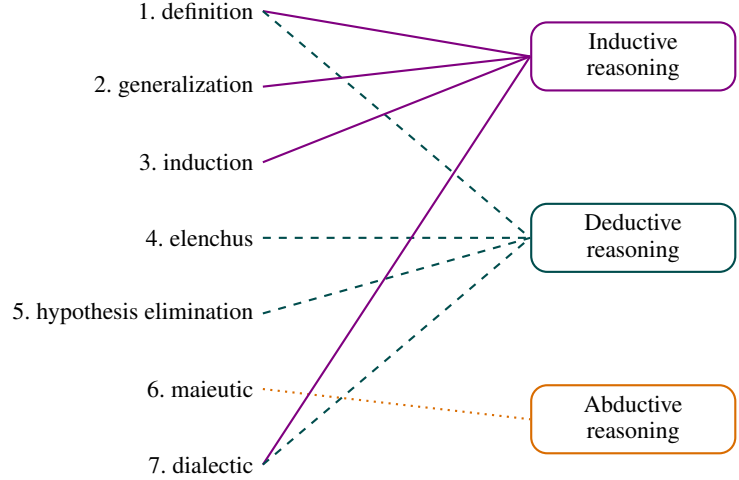
\begin{figure}[htbp]
\centering
\begin{tikzpicture}[
  method/.style={anchor=east, font=\small},
  family/.style={draw, rounded corners=6pt, thick, minimum width=2.7cm,
                 minimum height=0.9cm, align=center, font=\small},
  ind/.style={draw=violet, thick},
  ded/.style={draw=teal!60!black, thick, dashed},
  abd/.style={draw=orange!85!black, thick, dotted}
]
\node[method] (m1) at (0, 3.0) {1.\ definition};
\node[method] (m2) at (0, 2.0) {2.\ generalization};
\node[method] (m3) at (0, 1.0) {3.\ induction};
\node[method] (m4) at (0, 0.0) {4.\ elenchus};
\node[method] (m5) at (0,-1.0) {5.\ hypothesis elimination};
\node[method] (m6) at (0,-2.0) {6.\ maieutic};
\node[method] (m7) at (0,-3.0) {7.\ dialectic};
\node[family, draw=violet]           (find) at (4.9, 2.4) {Inductive\\reasoning};
\node[family, draw=teal!60!black]    (fded) at (4.9, 0.0) {Deductive\\reasoning};
\node[family, draw=orange!85!black]  (fabd) at (4.9,-2.4) {Abductive\\reasoning};
\draw[ind] (m1.east) -- (find.west);
\draw[ded] (m1.east) -- (fded.west);
\draw[ind] (m2.east) -- (find.west);
\draw[ind] (m3.east) -- (find.west);
\draw[ded] (m4.east) -- (fded.west);
\draw[ded] (m5.east) -- (fded.west);
\draw[abd] (m6.east) -- (fabd.west);
\draw[ind] (m7.east) -- (find.west);
\draw[ded] (m7.east) -- (fded.west);
\end{tikzpicture}
\caption{The Socratic methods and the IDA mapping: each of the seven
methods drawn to the classical family or families it exercises (solid,
inductive; dashed, deductive; dotted, abductive). The view is deliberately
coarse: Table~\ref{tab:ocr-crit-roof} refines it with the defeasible,
legal-evidentiary, causal, and practical families, and a method can invoke
more than one family, as hypothesis elimination is deductive in form
(refutation) and abductive in function (narrowing the hypothesis set). No
method maps to a causal family: the tradition supplies testing procedures
but no causal type system, the gap Pearl's ladder fills in
Section~\ref{sec:ocr-pearl-map}.}
\label{fig:ocr-ida}
\end{figure}

The oldest layer of the tradition makes the same point procedurally. The
seven Socratic methods, definition, generalization, induction, elenchus,
hypothesis elimination, maieutic, and dialectic, are not opinions about
reasoning; they are testing procedures, and each maps onto the formal
families: definition and dialectic exercise inductive and deductive
reasoning together, generalization and induction are inductive, elenchus and
hypothesis elimination are deductive refutation, and maieutic questioning is
abductive (Figure~\ref{fig:ocr-ida}). The mapping also reveals what the
tradition lacks: it supplies testing procedures but no causal type system;
that gap waits two millennia for the ladder of
Section~\ref{sec:ocr-pearl-map}. CRIT operationalizes the tradition for
machine evaluation
(Section~\ref{sec:ocr-reasoning-vs-debate}), which is why it sits at the
roof of the constructive sections. The hammer question extends accordingly:
where, in a language model, is elenchus? A refutation procedure requires a
counter-position held and pressed against the claim, and a solitary model
sampling from adjacent conditionals holds none.

One more absence completes the critique, and it is social rather than
formal. Mercier and Sperber argue that human reasoning evolved not mainly
for solitary truth-seeking, but for producing and evaluating arguments in
social exchange~\cite{mercier2011why,mercier2017enigma}. The same lesson
applies to machine reasoning. A claim must be asked: What supports it? What
would defeat it? What evidence is missing? What alternative explanation has
not been considered? Evaluation is therefore a division of labor: one side
proposes a claim, and another side tests it. A solitary model, sampling from
closely related conditionals, contains no second side.

\paragraph{From critique to construction.} The three layers agree on what
is missing, and it is not more reasoning-like text. Every instrument
surveyed above elicits reasoning; none of them writes reasoning down in a
form that survives the episode. A chain of thought is discarded when the
answer is emitted; a debate transcript is prose, not data; a reward is a
scalar that remembers nothing about which step earned it. What the field
lacks is a shared, typed, versioned schema for recording what was claimed,
what kind of reasoning the claim required, what evidence and warrants were
offered, what was tested and how, what failed, what is missing, and what
verdict followed, with each field backed by the formal construct that
Table~\ref{tab:ocr-absence} names. Procedures such as chain-of-thought,
retrieval, self-critique, discussion, refutation, and debate remain useful,
but only insofar as they improve the typed support relation the record
captures. They are instruments; the reasoning relation is the object. The
next section defines the record.

\section{The TRACE Schema}
\label{sec:ocr-record}

\subsection{Terminology Definition}
\label{sec:ocr-terminology}

\begin{table}[H]
\centering
\footnotesize
\begin{tabular}{p{0.20\linewidth}p{0.70\linewidth}}
\toprule
\textbf{Term} & \textbf{Definition} \\
\midrule
TRACE & Typed Reasoning And Commitment Evidence: the contract as a
whole, comprising the TRACE schema, the TRACE writer contract, and the
TRACE discipline. \\
\addlinespace[2pt]
TRACE schema & The versioned type definition, frozen as v1.0 before any
stage is implemented, specifying the record types, their fields, and the
fields' semantics. \\
\addlinespace[2pt]
TRACE record & One instance conforming to the TRACE schema; the root type
is \texttt{TraceRecord}. After this section, ``the record'' always denotes
a TRACE record. \\
\addlinespace[2pt]
TRACE causal record & The causal specialization \texttt{TraceCausalRecord},
adding rung, identification status, bounds, and missing evidence. \\
\addlinespace[2pt]
TRACE writer & Any process that emits a valid TRACE record; the eight-stage
procedure of Section~\ref{sec:ocr-trace-method} is the reference writer. \\
\addlinespace[2pt]
TRACE policy & The declared, versioned gates, weights, and thresholds under
which a record's score vector and verdict are computed; named by the
record's \texttt{policy\_version} field. \\
\addlinespace[2pt]
TRACE \newline discipline & No durable state change without a TRACE record carrying
a verdict. \\
\addlinespace[2pt]
TRACE-Bench & The benchmark and ablation protocol that judges schema
expressiveness, writer quality, and consumer value. \\
\bottomrule
\end{tabular}
\caption{The complete TRACE vocabulary. Eight terms, no aliases, every term
tied to TRACE.}
\label{tab:ocr-terms}
\end{table}

\subsection{Schema Definition and Key Fields}
\label{sec:ocr-schema-fields}

The TRACE schema is a versioned JSON type definition, stated in the vocabulary of Table~\ref{tab:ocr-terms}. Version 1.0 defines
two types: the root \texttt{TraceRecord} and its causal specialization
\texttt{TraceCausalRecord}. The complete type listings appear in
Appendix~\ref{app:schema}; the fields that carry the paper's arguments are:

\begin{itemize}
    \item \emph{Typing fields}: \texttt{claim\_text}, \texttt{claim\_type},
    \texttt{reasoning\_family}, \texttt{argument\_scheme}. They fix the
    standard a claim is tested against
    (Sections~\ref{sec:ocr-reasoning-vs-debate}
    and~\ref{sec:ocr-reasoning-families}).
    \item \emph{Anatomy fields}: \texttt{premises}, \texttt{warrants},
    \texttt{role\_states}, \texttt{defeaters}. They make hidden premises
    visible, including reconstructed enthymemes and their charity delta
    (Section~\ref{sec:ocr-reasoning-families}).
    \item \emph{Evidence and settlement fields}: \texttt{evidence},
    \texttt{source\_credibility}, \texttt{criticisms},
    \texttt{attack\_graph}. They carry audited support and the accounting
    of objections (Section~\ref{sec:ocr-reasoning-families}).
    \item \emph{Causal fields}, in \texttt{TraceCausalRecord}:
    \texttt{causal\_rung}, \texttt{graph\_fragment},
    \texttt{identifiability\_status}, \texttt{bounds},
    \texttt{missing\_evidence} (Section~\ref{sec:ocr-pearl-map}).
    \item \emph{Outcome fields}: \texttt{final\_status},
    \texttt{score\_vector}, \texttt{failed\_gates}, \texttt{missing},
    \texttt{repair}, \texttt{cost} (Section~\ref{sec:ocr-measurement}).
    \item \emph{Infrastructure fields}: \texttt{schema\_version},
    \texttt{writer\_id}, \texttt{policy\_version}, \texttt{provenance},
    \texttt{revision\_history}, \texttt{regret\_score},
    \texttt{consumer\_actions} (Sections~\ref{sec:ocr-measurement}
    and~\ref{sec:ocr-consumers}).
\end{itemize}

Three rules govern the schema. First, outcomes are fields: the verdict,
score vector, failed gates, missing items, repair, and cost live inside the
record, not beside it; a verdict separated from the record that justifies it
cannot exist, and every consumer reads one artifact. Second, versioning:
fields are added by version and deprecated by version, never silently
mutated; every consumer declares the schema versions it can read; freezing
version 1.0 is the first phase of the implementation plan. Third,
judgment: a schema cannot be optimal, only adequate, and adequacy is
measured on three axes, expressiveness (can the record represent rung
collapse, an implicit but essential warrant, a defer that names its missing
evidence), cost (reported as CostNorm in the metric vector), and consumer
value (the admission, plan-gating, and closed-loop experiments of the next
implementation plan). None of these axes requires an objective scalar of reasoning
quality.

\subsection{The TRACE Writer}
\label{sec:ocr-trace-method}

The TRACE writer is an API. Its signature:

\begin{quote}
\small
\begin{verbatim}
write(record: TraceRecord) -> WriteResult

WriteResult = {
  record_id,
  schema_version,
  policy_version,
  validation: pass | fail(violations),
  stored: true | false
}
\end{verbatim}
\end{quote}

The input is a JSON document declaring conformance to a TRACE schema
version. The service validates the document against that version; on pass,
the record is parsed and stored in the TRACE repository under the TRACE
schema; on fail, the write is rejected with the violated constraints named,
and nothing is stored. Writes are atomic and the repository is append-only:
a record is never mutated in place, and corrections enter through
\texttt{revision\_history}.

Who constructs the record is deliberately outside the API. Any process that
emits a valid record is a conforming writer: a caller can construct its
own, a single structured model call can emit one whole, and the reference
writer, an eight-stage procedure that builds a record from raw text under
per-stage contracts, is specified in Appendix~\ref{app:writer} with its algorithms,
stage table, modality contract, and cost model. TRACE-Bench compares the
reference writer against the single-pass writer as the staged
decomposition's null hypothesis (Section~\ref{sec:ocr-tracebench}).

One objection should be met head-on rather than left implicit. The reference
writer's model-mediated stages use the very mechanism whose limits
Section~\ref{sec:ocr-why-reasoning} argues, so the specification may appear
circular: the associative generator audits itself. The answer is
architectural. The schema converts an unreliable \emph{generator} into a
checkable \emph{proposer}: every model-mediated emission lands in a typed
field that validation, gates, perturbations, and the human-agreement boundary
can check, and nothing a stage emits becomes durable except through those
checks. The specification does not assume the stages execute reliably; it
requires their reliability to be measured at the typing boundary, and the
falsification criteria of Section~\ref{sec:ocr-falsify} already stake the
pipeline's warrant on that measurement.

\subsection{The Record-Consumer Contract}
\label{sec:ocr-consumer-contract}

The writer produces records; consumers, memory admission, plan gating,
temporal regret, and verdict reuse, act on them. Because a record is the only
channel between reasoning and durable action, the relationship between a
record and a consumer is itself a contract, and stating it is part of the
specification. The \texttt{write} API above, and the \texttt{consume} API of
the companion implementation, give the call signatures; this subsection states
the invariants those calls must preserve, which a signature alone does not
capture. The contract has two sides.

The record guarantees, to any conforming consumer, four properties.
\emph{Immutability}: a stored record is never mutated in place, so a consumer
that has read record version $v$ may rely on $v$ remaining exactly as read;
corrections arrive as new revisions linked through \texttt{revision\_history},
never as silent edits. \emph{Version legibility}: every record names the
schema and policy versions under which it was written, so a consumer can
refuse a record whose versions it does not understand rather than
misinterpret it. \emph{Self-containment}: the verdict, failed gates, missing
items, and repair live inside the record, so a consumer never needs the
writer's private state to act. \emph{Provenance}: every field is attributable
to the stage and inputs that produced it, so a consumer can audit why a
verdict holds.

The consumer, in return, accepts four obligations. \emph{Version pinning}: a
consumer declares which schema and policy versions it reads and rejects
records outside that set rather than coercing them. \emph{Fail closed}: an
incomplete, invalid, or unreadable record, or one whose decisive evidence is
unavailable, must never be treated as acceptance; the safe default is to
withhold the durable action. \emph{Authority separation}: the record's
verdict is advisory about \emph{warrant}, not a grant of \emph{authority};
the consumer owns the mapping from verdict to durable action under its own
declared, versioned policy, and a consumer may hold a technically acceptable
claim when its own stakes demand more, or escalate to a human. \emph{Action
write-back}: a consumer that acts on a record appends its decision to
\texttt{consumer\_actions}, so the record accumulates its own downstream
consequences and the loop from reasoning to action to outcome stays auditable
end to end.

Two corollaries follow that the flood vignette
(Section~\ref{sec:ocr-flood-vignette}) makes concrete. First, a TRACE verdict
and a consumer decision are different objects: \texttt{defer} is a property of
the record, while \textsc{Hold} is a plan-gate action the consumer writes back
after reading it. Second, the same record may license different actions for
different consumers, because authority separation lets a high-stakes consumer
withhold what a low-stakes consumer would commit, without any change to the
record. The contract is what makes the schema an operational interface rather
than a passive document.

\section{The Reasoning Hierarchy}
\label{sec:ocr-hierarchy}

The schema's typing fields promise that every claim is tested at the
standard its type demands. This section builds the hierarchy that redeems
the promise: the Socratic roof, the four evaluation layers, the reasoning
families with their argumentation toolbox, and Pearl's ladder as the causal
type system. The hierarchy has a cast, and Table~\ref{tab:ocr-cast}
introduces it at once: each figure, what he contributed, and which part of
the TRACE contract his machinery became. The scholars who contest the
lists, Wellman, Govier, Pollock, Gordon, Freeman, and Hitchcock, appear in
Appendix~\ref{app:anatomy}, where the contested extensions live.

\begin{table}[H]
\centering
\footnotesize
\begin{tabular}{p{0.16\linewidth}p{0.36\linewidth}p{0.26\linewidth}p{0.10\linewidth}}
\toprule
\textbf{Figure} & \textbf{Contribution} & \textbf{What TRACE takes} & \textbf{Where} \\
\midrule
Socrates & the testing procedures: definition, generalization, induction, elenchus, hypothesis elimination, maieutic, dialectic & the questioning stages, operationalized by CRIT & Section~\ref{sec:ocr-reasoning-vs-debate} \\
\addlinespace[2pt]
Toulmin~\cite{toulmin1958uses} & the anatomy of a single argument: claim, grounds, warrant, backing, qualifier, rebuttal & the anatomy fields and the \texttt{role\_states} enum & Section~\ref{sec:ocr-reasoning-families} \\
\addlinespace[2pt]
Walton~\cite{walton2008schemes} & argument schemes, each owing its critical questions & the scheme registry and the stage-3 question batteries & Section~\ref{sec:ocr-reasoning-families} \\
\addlinespace[2pt]
Dung~\cite{dung1995acceptability} & abstract argumentation: attack graphs and acceptability semantics & the \texttt{attack\_graph} field and the settlement rule & Section~\ref{sec:ocr-reasoning-families} \\
\addlinespace[2pt]
Pearl~\cite{pearl2009causality} & the causal hierarchy: association, intervention, counterfactuals, with do-calculus and identification & the causal fields: rung, identifiability status, bounds & Section~\ref{sec:ocr-pearl-map} \\
\bottomrule
\end{tabular}
\caption{The hierarchy's cast: five bodies of theory and the TRACE
machinery each became. The contested extensions and their proponents are
in Appendix~\ref{app:anatomy}.}
\label{tab:ocr-cast}
\end{table}

\subsection{Reasoning Before Debate: The CRIT Roof}
\label{sec:ocr-reasoning-vs-debate}

Section~\ref{sec:ocr-problem} separated reasoning from the facilities that
may elicit or test it. This section uses debate as the clearest stress case
for that distinction. Debate is the most explicit adversarial reasoning facility:
it assigns roles, creates opposition, elicits counter-reasons, and makes
disagreement visible. But the same point applies to chain-of-thought,
retrieval, self-critique, discussion, and refutation. None of these facilities
defines reasoning. Each is useful only insofar as it improves the support relation linking a
claim to its reasons, evidence, warrants, counter-reasons, and context.

Reasoning and debate therefore answer different questions. \emph{Reasoning}
asks whether a conclusion is supported by premises under a valid or acceptable
inference pattern. \emph{Debate} asks whether competing agents or roles can
expose missing premises, alternative explanations, contradictions, and
overclaims. Debate can improve reasoning, but it does not define reasoning.
Before debate can be useful, the system must know what kind of claim is being
made and what standard of support that claim requires.

CRIT is the starting point for this volume's architecture because it supplies
a facility-independent evaluator.
Given a document \(d\), CRIT identifies the conclusion
\(\Omega\), extracts supporting reasons \(R\), evaluates each
reason-to-conclusion relation \(r \Rightarrow \Omega\), recursively audits
reasons that are themselves claims, seeks rival reasons \(R'\), evaluates
counter-reasons, aggregates validity and credibility scores, and then asks for
reflective synthesis in context~\cite{chang2023crit}. The public CRIT paper
explicitly develops definition, elenchus, dialectic, maieutics,
generalization, and counterfactual reasoning as prompt-template methods
connected to inductive, deductive, and abductive reasoning~\cite{chang2023crit}.

This gives the paper its roof:

\begin{enumerate}[leftmargin=1.7em]
  \item \textbf{Claim first.} What is the conclusion, and what kind of conclusion is it?
  \item \textbf{Reason next.} What reasons are offered, and what inference relation connects them to the conclusion?
  \item \textbf{Evidence and credibility.} What evidence supports each reason, and how credible are the sources?
  \item \textbf{Rival reasons.} What counterarguments or alternative explanations have not been considered?
  \item \textbf{Context.} Would the support relation change under a different time, place, population, legal regime, institutional role, or value frame?
  \item \textbf{Status.} Should the claim be accepted, accepted with qualification, revised, deferred, or rejected?
\end{enumerate}

\begin{table}[htbp]
\centering
\footnotesize
\begin{tabular}{p{0.16\linewidth}p{0.15\linewidth}p{0.18\linewidth}p{0.39\linewidth}}
\toprule
\textbf{CRIT method} & \textbf{Evaluator role} & \textbf{Families invoked} & \textbf{Question it puts to the argument} \\
\midrule
Definition & Claim and term typing & Deductive, definitional & What exactly is claimed, and what terms and scope conditions fix its meaning? \\
\addlinespace[2pt]
Elenchus (cross-examination) & Support testing & Deductive, defeasible, legal & Does each reason support the conclusion, or does it hide a contradiction, missing premise, weak warrant, or unsupported source? \\
\addlinespace[2pt]
Evidence typing & Evidentiary classification & Inductive, statistical, legal & Is the support a theory, an opinion, a statistic, testimony, or a cited claim, and does it have the evidentiary status the claim requires? \\
\addlinespace[2pt]
Source audit (recursive) & Evidentiary tracing & Defeasible, legal, probabilistic & If a reason is itself a claim, what supports that subclaim, and how credible is its source chain? \\
\addlinespace[2pt]
Dialectic & Rival-reason search & Abductive, defeasible, dialectical & What counterarguments, alternative explanations, or omitted evidence would defeat or weaken the claim? \\
\addlinespace[2pt]
Hypothesis elimination & Explanation pruning & Deductive plus abductive & Which candidate explanations fail on contradictions, failed predictions, or missing evidence? \\
\addlinespace[2pt]
Generalization & Instance-to-rule check & Inductive, analogical & Does the move from instances to a broader rule respect sample and similarity conditions? \\
\addlinespace[2pt]
Maieutics & Reflective articulation & Reflective, meta-cognitive & Can the evaluator explain why the score was assigned, and what repair would strengthen the claim? \\
\addlinespace[2pt]
Counterfactual questioning & Context shift and robustness & Causal, abductive, practical & Would the support relation survive a change of time, place, population, action, or background conditions? \\
\bottomrule
\end{tabular}
\caption{The CRIT roof in one table: each Socratic or legal-evidentiary method, the evaluator role it plays, the reasoning families it invokes, and the question it asks. Chain-of-thought, retrieval, self-critique, refutation, discussion, or debate
may instantiate some of these operations, but the operations themselves
evaluate reasoning.}
\label{tab:ocr-crit-roof}
\end{table}

The legal flavor of CRIT is not accidental. Elenchus is cross-examination: it tests whether the reason actually supports the conclusion, whether the support rests on evidence or assertion, whether the source is credible, and whether a reason is itself an unsupported claim. Dialectic introduces adverse positions and counter-reasons. Recursive validation resembles evidentiary tracing: when a premise is a quoted claim, the evaluator follows the source chain. These are legal-evidentiary techniques in computational form, even when the domain is not law. Table~\ref{tab:ocr-crit-roof} assembles the full method set: for each CRIT method, the evaluator role it plays, the reasoning families it invokes, and the question it puts to the argument.

Table~\ref{tab:ocr-crit-roof} also fixes a boundary. Chain-of-thought,
retrieval, self-critique, discussion, refutation, SocraSynth, and EVINCE may
instantiate some of these operations, and SocraSynth or EVINCE may organize or
measure them in debate. But CRIT defines what all of them are trying to
improve: the reason-to-claim support relation.

The third column of Table~\ref{tab:ocr-crit-roof} deserves emphasis. CRIT's Socratic methods are not another taxonomy competing with deduction, induction, and abduction; they are evaluator operations that invoke those families. Definition fixes terms before any inference is scored. Elenchus tests deductive and defeasible support. Evidence typing and source audit carry the inductive and legal-evidentiary load. Dialectic and hypothesis elimination drive abductive comparison. Generalization checks inductive strength, and counterfactual questioning probes causal and practical robustness.

This mapping is the roof under the rest of the paper. It also explains why causal reasoning cannot be treated as the whole of reasoning. A causal claim may require deductive graph reasoning, inductive estimation, abductive diagnosis, and counterfactual evaluation. Those operations play different roles and should not be collapsed into one score.

\subsection{Four Layers of Typed Evaluation}
\label{sec:ocr-four-layers}

Table~\ref{tab:ocr-relatedwork-layers} situates the related work by the layer it serves.

The typed-reasoning space divides into four layers. Each layer answers a different evaluator question.

\begin{enumerate}[leftmargin=1.7em]
  \item \textbf{Claim types:} factual, definitional, causal, counterfactual, predictive, diagnostic, normative, legal, policy, preference, explanatory.
  \item \textbf{Reasoning types:} deductive, inductive, abductive, analogical, causal, probabilistic, defeasible, practical, deontic, dialectical, narrative, contextual.
  \item \textbf{Argument schemes:} expert opinion, witness testimony, sign, cause--effect, correlation--cause, analogy, precedent, rule, consequences, practical reasoning, values, ignorance, popular opinion, slippery slope, threat, commitment.
  \item \textbf{Context and value evaluation:} audience, stakes, vulnerability, culture, role, institutional norm, fairness, harm, autonomy, reversibility, proportionality.
\end{enumerate}

The second and third layers are easy to conflate, and the distinction matters: a reasoning family names the formal standard of inference, while an argument scheme names a recurring warrant pattern in discourse. A causal claim may instantiate a correlation-to-cause scheme, but the causal family supplies the rung and identification tests. The scheme tells the evaluator which critical questions to ask; the family tells it which formal standard the answers must meet.

The literature base supports these layers but does not by itself provide an agentic evaluation architecture. Argumentation schemes classify defeasible reasoning moves and associate each move with critical questions~\cite{walton2008schemes,walton2006fundamentals,walton2013methods}. Computational argumentation formalizes attacks, defenses, and acceptability among competing arguments~\cite{dung1995acceptability,prakken2002logics,baroni2011semantics,benchcapon2007argumentation}, with structured frameworks supplying the internal anatomy of premises, rules, and attacks~\cite{prakken2010structured,modgil2014aspic,toni2014aba}. Value-based and practical-reasoning accounts show why the same evidence can support different conclusions under different goals, audiences, and values~\cite{benchcapon2003value,atkinson2007practical}. Argument mining and argument-quality research provide computational tools for extracting and assessing claims, premises, support, attack, and quality dimensions~\cite{lippi2016argumentation,lawrence2019argument,wachsmuth2017quality,wachsmuth2024argument,li2025llmargumentmining}. Context-aware reasoning motivates DIKE--ERIS: argument relevance depends on task, role, stakeholder, situation, and institutional setting~\cite{dey2001context,perera2014context,du2024contextaware}. Surveys of reasoning with language models map the modern elicitation and evaluation landscape~\cite{qiao2022reasoning,sun2023foundationreasoning,liu2025logical,huang2023towards}, and causal inference disciplines embedded causal subclaims~\cite{pearl2009causality,imbens2015causal,peters2017elements,guo2020survey,feder2022causal}.

Four further traditions bear directly on the evaluator. First, deductive reasoning has a mature verification stack of its own. Automated theorem proving and proof assistants such as Lean give the deductive family a machine-checkable standard~\cite{demoura2021lean}; benchmarks now bridge informal and formal proof~\cite{welleck2021naturalproofs,zheng2022minif2f,jiang2023draft}; and trained verifiers with process supervision bring stepwise checking to natural-language reasoning~\cite{cobbe2021verifiers,lightman2024verify}. When the family typing of Section~\ref{sec:ocr-reasoning-families} says deductive, these are the plug-in checkers. Second, belief revision and truth maintenance supply the formal account of how a TRACE record must change when a premise falls: AGM belief revision~\cite{agm1985}, truth-maintenance systems~\cite{doyle1979tms,dekleer1986atms}, and nonmonotonic logics~\cite{reiter1980default,mccarthy1980circumscription} are the classical machinery behind the record's revision history and RLER's policy updates. Third, decision theory prices the evaluator's choices: subjective expected utility~\cite{savage1954foundations}, applied statistical decision theory~\cite{raiffa1961applied}, and the value of information~\cite{howard1966voi} ground ERM's regret weights, the stopping rules of Section~\ref{sec:ocr-stopping}, and the defer verdict's obligation to name the evidence that would most reduce uncertainty. Fourth, the AI-and-law tradition formalizes the legal-evidentiary techniques CRIT borrows: case-based reasoning with precedents~\cite{ashley1990hypo}, dialectical assessment of conflicting legal arguments~\cite{prakken1996dialectical}, case comparison incorporating theories and values~\cite{benchcapon2003theories}, and computational models of burden of proof~\cite{gordon2007carneades}. These give the roof's cross-examination, source audit, and burden allocation their disciplinary home.

One prior artifact deserves direct comparison because it is the closest in
kind. The Argument Interchange Format (AIF) is the established schema for
representing and exchanging arguments as typed graphs of information and
scheme nodes, with an ecosystem of corpora and tools built on
it~\cite{chesnevar2006aif,lawrence2012aifdb}. TRACE shares AIF's premise,
that arguments deserve a typed, machine-readable representation, and departs
from it on exactly the dimensions an operational consumer needs: AIF records
an argument's structure, while TRACE additionally records its
\emph{adjudication}. The verdict, score vector, failed gates, missing items,
and repair are fields of the record rather than annotations beside it;
reasoning families route subclaims to formal plug-ins with rung and
identification status for causal claims; the schema is versioned with an
append-only revision history; and the record-consumer contract of
Section~\ref{sec:ocr-consumer-contract} states what a downstream system may
rely on. AIF is an interchange format for arguments; TRACE is an interchange
format for \emph{evaluated} arguments and the commitments they license.

The design gap is now precise. Existing work gives taxonomies, dialogue semantics, extraction tools, causal methods, and context theories. What the paper supplies is the evaluation-side organization an agentic system needs: which quality standard applies to which reasoning family, how a mixed argument is decomposed and routed, how causal subclaims are mapped to Pearl's rungs, and how final judgments are scored, audited, and revised.

\begin{table}[H]
\centering
\small
\begin{tabular}{p{0.20\linewidth}p{0.33\linewidth}p{0.37\linewidth}}
\toprule
\textbf{Layer} & \textbf{Representative literature} & \textbf{Role in this paper} \\
\midrule
Claim type & CRIT, Toulmin, causal inference & Decide what kind of conclusion is being asserted. \\
\addlinespace[2pt]
Reasoning family & Logic, statistics, abduction, analogy, Pearl, decision theory & Select the standard: validity, calibration, explanation, identification, utility, obligation. \\
\addlinespace[2pt]
Argument scheme & Walton schemes, computational argumentation, argument mining & Identify warrants, critical questions, attacks, rebuttals, and defeaters. \\
\addlinespace[2pt]
Context and value & Value-based argumentation, context-aware computing, DIKE--ERIS & Decide whether the argument is appropriate under audience, stakes, institution, values, and harm. \\
\bottomrule
\end{tabular}
\caption{How related work is integrated into the four-layer evaluation architecture.}
\label{tab:ocr-relatedwork-layers}
\end{table}

\subsection{Families of Reasoning}
\label{sec:ocr-reasoning-families}

Once CRIT identifies the claim and reasons, the evaluator must classify the reasoning family. A single passage may contain several families at once. For example, a medical recommendation may contain a diagnostic inference, a causal treatment claim, a probabilistic risk estimate, a practical-action argument, and a value judgment about patient goals. Treating all of them as ``causal'' or all of them as ``debate'' loses the type-specific standards. Table~\ref{tab:ocr-reasoning-families} defines the ten families this paper uses: the surface signature by which each is recognized in text, the standard it must meet, and its typical failure.

\begin{table}[htbp]
\centering
\footnotesize
\begin{tabular}{p{0.15\linewidth}p{0.34\linewidth}p{0.20\linewidth}p{0.17\linewidth}}
\toprule
\textbf{Family} & \textbf{Surface signature} & \textbf{Primary standard} & \textbf{Typical failure} \\
\midrule
Deductive & necessity modals (\emph{must}, \emph{cannot}), universal quantifiers, definitions & validity, soundness & invalid form; false premise \\
\addlinespace[2pt]
Inductive/ statistical & sample-to-general moves (\emph{in most cases}, \emph{studies show}), rates & sample adequacy, reference class, calibration & biased sample; base-rate neglect \\
\addlinespace[2pt]
Abductive & \emph{best explains}, \emph{likely cause}, diagnosis verbs & coverage, simplicity, contrast with rivals & just-so story; single-hypothesis fixation \\
\addlinespace[2pt]
Analogical/ case-based & \emph{like}, \emph{just as}, precedent invocation & relational correspondence, disanalogy check & surface similarity \\
\addlinespace[2pt]
Causal & rung markers: \emph{is associated with} (1); \emph{if we do} (2); \emph{would have, had we} (3) & rung discipline, identification, assumptions & rung collapse; confounding \\
\addlinespace[2pt]
Probabilistic & credences, odds, \emph{likely}/\emph{unlikely}, hedges & coherence, likelihood, calibration & overconfidence; ignored dependence \\
\addlinespace[2pt]
Defeasible & \emph{normally}, \emph{unless}, \emph{as a rule} & critical questions, defeat handling & ignored exceptions; unresolved defeater \\
\addlinespace[2pt]
Practical & \emph{should}, \emph{the right course}, means--end verbs & means--end fit, alternatives, side effects, proportionality & suppressed warrant; no alternatives \\
\addlinespace[2pt]
Deontic/legal/ normative & \emph{must not}, \emph{duty}, \emph{right}, rule and precedent citations & rule fit, authority, burden, value transparency & is--ought slide; hidden value premise \\
\addlinespace[2pt]
Dialectical/ contextual & concession and reply markers (\emph{granted}, \emph{however}) & burden of proof, issue resolution & consensus mistaken for truth \\
\bottomrule
\end{tabular}
\caption{The ten reasoning families: the surface signature by which stage-zero
detection recognizes and routes each family, the quality standard it must
meet, and the failure its tests most often catch.}
\label{tab:ocr-reasoning-families}
\end{table}

\subsubsection{The Argumentation-Theory Toolbox: Anatomy, Schemes, Acceptability}
\label{sec:ocr-argtheory}
 
Argumentation theory supplies a complementary layer, and it is worth being
precise about what each of its three traditions contributes, because the
pipeline of this paper uses them for three different jobs. Toulmin
supplies the \emph{anatomy} of a single argument; Walton supplies the
\emph{tests} for a single inference pattern; abstract argumentation supplies
the \emph{accounting} when many arguments attack one another. None of the
three replaces CRIT; CRIT is the operator that runs all three.
 
\paragraph{Toulmin: the anatomy of one argument.} Toulmin's model
reconstructs any single-step argument into six roles~\cite{toulmin1958uses}:
the \emph{claim} (what is being asserted), the \emph{grounds} (the data
offered in support), the \emph{warrant} (the usually unstated bridge that
licenses the step from grounds to claim), the \emph{backing} (what supports
the warrant itself), the \emph{qualifier} (the claimed strength:
\emph{certainly}, \emph{presumably}, \emph{some}, \emph{all}), and the
\emph{rebuttal} (the conditions under which the claim would fail). Toulmin's
own example carries all six in one sentence: \emph{Harry was born in Bermuda}
(grounds), \emph{so, presumably} (qualifier), \emph{Harry is a British
subject} (claim), \emph{since a man born in Bermuda will generally be a
British subject} (warrant), \emph{on account of the relevant statutes}
(backing), \emph{unless both his parents were aliens} (rebuttal). Two
properties of the layout matter as much as the roles themselves. First, the
reconstruction is not unique: the boundary between grounds and warrant shifts
with the analyst, which is why the pipeline records the reconstruction it
chose and why the charity delta measures
that choice's distance from the argument as stated. Second, roles may be
empty: a deductively valid argument has no substantive qualifier or rebuttal,
and an evaluable record marks such roles absent rather than inventing
content for them. The anatomy matters operationally because most real
arguments arrive with only the claim and the grounds visible; the warrant is
hidden, the qualifier is inflated, and the rebuttal is missing. Three of
CRIT's core moves are exactly the recovery of the missing roles: warrant
explicitness surfaces the bridge, quantifier consistency compares the
qualifier against the grounds, and rival-reason search asks for the rebuttal
the author omitted. The Toulmin roles are also where the
\texttt{TraceRecord} of Section~\ref{sec:ocr-record} gets its fields;
Table~\ref{tab:ocr-extended-anatomy} maps each role to the record field and
pipeline component that handles it.

\paragraph{Is the six-role list complete, and does it matter?} No
completeness theorem exists at the level of natural argument, and the
literature contests the list in both directions, proposing additions such
as the counter-consideration, the undercutting defeater, and presumption
with burden of proof, and proposing merges in the other
direction~\cite{wellman1971challenge,govier1987problems,pollock1987defeasible,gordon2007carneades,freeman1991dialectics,hitchcock2006arguing}.
Two answers make the schema robust to this churn. The first is formal:
in the structured-argumentation core used here, one argument step is
premises, rule, and conclusion, so its attack surfaces are exactly three,
premise, rule, and conclusion, and closure holds by
construction~\cite{prakken2010structured,modgil2014aspic}; Toulmin's six
roles are a readable projection of that closed core, and every proposed
extension already has a home, counter-considerations and undercutters in
the attack graph, presumption and burden in external adjudication. The
second is architectural: the schema does not adjudicate taxonomy disputes,
it versions them. The role vocabulary, the scheme registry, the family
taxonomy, and the gate metrics are all versioned under the TRACE policy, so
adding a role, subtracting a scheme, or admitting a future theory's family
and its critical-question battery is a version bump, not a refutation; the
contract absorbs theoretical progress the way durable interchange standards
do, by extension rather than by breakage. Appendix~\ref{app:anatomy} carries the full
two-tier development and the extended anatomy table
(Table~\ref{tab:ocr-extended-anatomy}).

\paragraph{Role states and reconstruction.} Number the roles $R_1$ claim
through $R_6$ rebuttal, with extensions $R_7$ counter-considerations, $R_8$
undercutters, and $R_9$ presumption and burden. For a given segment, each
role is in exactly one of five states, the values of the record's
\texttt{role\_states} field: \emph{stated}, \emph{elicited} (supplied by
the source in answer to a query), \emph{reconstructed} (supplied by the
evaluator and marked as such), \emph{absent}, or \emph{contradicted}
(the source's own statements are mutually inconsistent on this role). Reconstruction is a
procedure, not an art: roles are filled in dependency order; the source is
queried when one is available, which is CRIT's elenchus in interactive
form; a missing warrant is abduced as the minimal bridging proposition, and
if no plausible bridge exists the step is rejected as a non sequitur, a
hard gate. The procedure emits four scores consumed by
Section~\ref{sec:ocr-measurement}: warrant explicitness,
critical-question coverage, quantifier consistency, and the charity delta
$\Delta$, the load-weighted share of reconstructed roles. Elicited roles
improve the argument; reconstructed roles improve only the evaluator's
reconstruction of it, and $\Delta$ preserves the difference. The full
algorithm, with its formulas and its human-audit control on reconstructed
warrants, is Appendix~\ref{app:writer}.

\paragraph{Walton: pre-packaged cross-examination.} Where Toulmin gives the
parts list, Walton, Reed, and Macagno give the catalog of \emph{inference
patterns}: some sixty presumptive argumentation schemes, each paired with
the critical questions that probe precisely where that pattern
fails~\cite{walton2008schemes,walton2006fundamentals,walton2013methods}. A
scheme is a defeasible template: it confers provisional support, not proof,
and each critical question names a way the presumption can be defeated. The
practical value for an automated evaluator is hard to overstate: once the
scheme is identified, the cross-examination comes pre-written
(Table~\ref{tab:ocr-schemes-cq}). The evaluator does not need to invent
objections; it needs to \emph{route} to the right question list and record
which questions were answered, which were dodged, and which defeat the
argument outright. That record is the scheme-fit and critical-question
coverage that Section~\ref{sec:ocr-measurement} scores. In the reconstruction
algorithm above, Walton schemes supply the question set used in the rebuttal,
undercutter, and counter-consideration steps; their answered-over-applicable
ratio is the critical-question coverage component of \(\mathrm{SchemeFit}\).

The correlation-to-cause questions of Table~\ref{tab:ocr-schemes-cq}
shadow the causal family's own tests: the third-factor question is the
confounder audit, the reverse-causation question is edge orientation, and
the survives-intervention question is identification itself. The resemblance
is the design, not a redundancy: a scheme's critical questions are the
cheap, discourse-level interface to a family's formal battery, and one rule
prevents double charging. Where a question has a formal counterpart, the
plug-in adjudicates and the record marks the question answered by it; where
none exists, which holds for most of the catalog (partial counterparts are
emerging, such as source-credibility models for expert opinion and
decision-theoretic checks for practical reasoning), the critical-question
answer is itself the verdict, held to the defeasible standard. In one
sentence: schemes screen, families adjudicate.
 
\begin{table}[htbp]
\centering
\small
\begin{tabular}{p{0.22\linewidth}p{0.68\linewidth}}
\toprule
\textbf{Scheme} & \textbf{Representative critical questions} \\
\midrule
Expert opinion & Is the source a genuine expert, in \emph{this} field? What exactly did the expert assert? Is the expert credible and unbiased? Do other experts agree? Is the assertion backed by evidence? \\
\addlinespace[2pt]
Correlation to cause & Is the correlation real or artifactual? Could a third factor produce both variables? Could causation run in reverse? Is there a plausible mechanism? Would the relation survive intervention? \\
\addlinespace[2pt]
Negative consequences & How likely is the harm? How severe? Does the action actually produce it? Are there less costly alternatives that avoid it? \\
\addlinespace[2pt]
Practical reasoning & Does the action achieve the goal? Are there alternative actions? What are the side effects? Do other goals conflict? Is the action feasible? \\
\addlinespace[2pt]
Analogy & Are the two cases similar in the respects that matter? Are there relevant differences? Is there a counter-analogy? \\
\bottomrule
\end{tabular}
\caption{Walton schemes ship with their own cross-examination. Identifying
the scheme makes critique systematic rather than improvised: the
critical-question list is the intake test battery for that inference pattern.
Where a family plug-in supplies a formal counterpart, the plug-in adjudicates
the answer; otherwise the critical-question answers themselves carry the
defeasible verdict.}
\label{tab:ocr-schemes-cq}
\end{table}
 
\paragraph{Dung and structured argumentation: the accounting.} A real
evaluation rarely ends with one argument. The dialectic stage produces
counterarguments; counterarguments attract rebuttals; rebuttals attract
undercuts. Abstract argumentation~\cite{dung1995acceptability} answers the
bookkeeping question this creates: given a set of arguments and an attack
relation among them, which arguments are ultimately \emph{acceptable}? The
core intuition fits in a sentence: an argument stands if every argument
attacking it is itself defeated by an argument that stands: innocent until
attacked, reinstated when its attacker falls. Different semantics make this
circle precise in different ways~\cite{baroni2011semantics}, and structured
frameworks such as ASPIC\textsuperscript{+} and assumption-based
argumentation extend the accounting inside arguments, distinguishing attacks
on a premise, on the inference rule, and on the conclusion
itself~\cite{prakken2010structured,modgil2014aspic,toni2014aba,prakken2002logics,benchcapon2007argumentation}.
In the pipeline, this layer is the formal backbone of external adjudication
(condition D4 of Section~\ref{sec:ocr-debateness}): after debate has
populated the attack graph and CRIT has audited each node, acceptability
semantics, not the debaters' consensus, determines which claims survive.
 
\paragraph{Division of labor, in one sentence.} Toulmin tells the evaluator
what to extract from one argument; Walton tells it what to ask of one inference pattern; Dung tells it how to settle a field of competing
arguments; and CRIT is the evaluator that performs the extraction, asks the
questions, and reports the settlement. These traditions do not replace CRIT; they give CRIT the scheme-specific questions needed after the claim has been typed. Section~\ref{sec:ocr-endtoend} shows all three at work on a single sentence.

\subsection{Pearl's Ladder Inside the Reasoning Roof}
\label{sec:ocr-pearl-map}

Table~\ref{tab:ocr-pearl-map} states the mapping this subsection develops: Pearl's ladder as a causal type system inside the broader roof.

Pearl's hierarchy separates three causal question types: association, intervention, and counterfactual~\cite{pearl2009causality,pearl2018book}. In this paper, the hierarchy is not the roof; it is the type system for causal claims inside the roof.

\begin{enumerate}[leftmargin=1.7em]
    \item \textbf{Rung 1: Association.} What is observed together? This includes correlations, conditional probabilities, classification, prediction, and observational patterns.
    \item \textbf{Rung 2: Intervention.} What would happen if an action were imposed? This requires a target such as \(do(X=x)\) and assumptions that distinguish doing from seeing.
    \item \textbf{Rung 3: Counterfactual.} What would have happened under an alternative condition, given what actually occurred? This requires an individual or case-conditioned model, not merely a population association.
\end{enumerate}

The mapping to reasoning families is as follows.

\begin{table}[H]
\centering
\small
\begin{tabular}{p{0.16\linewidth}p{0.24\linewidth}p{0.25\linewidth}p{0.25\linewidth}}
\toprule
\textbf{Pearl rung} & \textbf{Primary reasoning families} & \textbf{Formal value} & \textbf{Boundary} \\
\midrule
Association & Inductive, statistical, probabilistic & Prediction, screening, pattern discovery, priors, hypothesis generation, calibration & Cannot by itself authorize an intervention. \\
\addlinespace[2pt]
Intervention & Causal, deductive-within-model, statistical identification & Estimates the effect of doing, policy choice, treatment selection, manipulation & Requires a causal model and identification assumptions. \\
\addlinespace[2pt]
Counterfactual & Causal, abductive, probabilistic, deductive-within-SCM & Individualized explanation, responsibility, regret, what-if learning, alternative-outcome analysis & Requires observed facts plus a structural model; cannot be obtained from association alone. \\
\bottomrule
\end{tabular}
\caption{Pearl's ladder as a causal type system inside the broader reasoning roof. Association is valuable formal reasoning, but it supports prediction and hypothesis formation rather than action by itself.}
\label{tab:ocr-pearl-map}
\end{table}

This answers three common confusions.

\paragraph{Is deductive reasoning causal?} No. Deduction is an inference relation: if the premises are true and the form is valid, the conclusion must be true. Causality is semantic content about how variables, events, or actions affect one another. A causal graph plus assumptions may allow some causal conclusions to be deduced, but that does not make deduction identical to causality. Deduction tests validity; causal reasoning tests intervention, identification, counterfactual dependence, and assumptions.

\paragraph{What is counterfactual reasoning?} Counterfactual reasoning asks about an alternative world conditioned on the actual world. In Pearl's structural account, it combines observed evidence with a model and an intervention-like modification. Informally, it often requires abduction to infer the case-specific background, action to modify the antecedent, and prediction to compute the consequent~\cite{pearl2009causality}. In CRIT terms, counterfactual questioning tests robustness under context shift; in ERM/RLER terms, it is also how the system asks what mistake would have been avoided under a different reasoning policy.

\paragraph{What is the value of association?} Association is not ``mere'' reasoning. It is essential for detecting regularities, making predictions, selecting candidate hypotheses, estimating priors, calibrating confidence, and deciding what evidence to collect next. Its limitation is precise: association cannot by itself support claims about what would happen under intervention or what would have happened otherwise. A typed evaluator should preserve association's value while blocking rung collapse.

For interventional claims, once a causal graph and evidence set are declared, the target effect may be identifiable, partially identifiable, or underidentified. If the available evidence cannot support the requested causal quantity, the correct output is not a confident answer. It is a bounded statement: the claim is underidentified, the available bounds are reported when possible, and the missing evidence needed to tighten the claim is named~\cite{shpitser2006identification,manski1990nonparametric,balke1997bounds}. This turns defer from a conversational refusal into a technical outcome.

\begin{proposition}[Rung-boundedness of debate]
\label{prop:ocr-rungbound}
Relative to a declared structural-model class and assumption set, if every
unit in the evidence ledger is rung-1 (observational) and the claim under
debate is rung-2 or rung-3, then no debate protocol, of any length
and among any number of agents, can warrant a verdict tighter than the
partial-identification bounds that ledger licenses
\cite{manski1990nonparametric,balke1997bounds,shpitser2006identification}.
The optimal reachable verdict is the bound, or defer. Debate can add
evidence, and it can surface assumptions which, once granted, change the
ledger's rung composition; but exchange alone cannot climb the ladder.
\end{proposition}

Proposition~\ref{prop:ocr-rungbound} is the point at which the paper's two
directions meet: \emph{Pearl's hierarchy bounds what debate can achieve;
debate determines whether the bound is achieved.} The semantic axis sets the
ceiling; the procedural axis determines the distance to it.

\section{The Writer Stages}
\label{sec:ocr-stages}

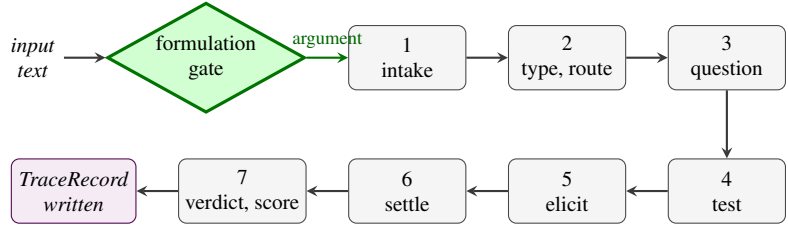
\begin{figure}[htbp]
\centering
\begin{tikzpicture}[
  box/.style={draw=gray!55!black, fill=gray!8, rounded corners=3pt,
              minimum height=8.5mm, minimum width=15.5mm, align=center,
              font=\footnotesize, inner sep=2.5pt},
  gate/.style={draw=green!45!black, fill=green!18, very thick, diamond,
               aspect=1.8, align=center, font=\footnotesize,
               inner sep=1.2pt},
  rec/.style={draw=violet!60!black, fill=violet!8, rounded corners=3pt,
              minimum height=8.5mm, align=center,
              font=\footnotesize\itshape, inner sep=3pt},
  io/.style={align=center, font=\footnotesize\itshape},
  arr/.style={->, >=stealth, thick, draw=gray!45!black},
  node distance=5mm and 5.5mm
]
\node[io] (in) {input\\text};
\node[gate, right=of in] (g) {formulation\\gate};
\node[box, right=of g] (s1) {1\\intake};
\node[box, right=of s1] (s2) {2\\type, route};
\node[box, right=of s2] (s3) {3\\question};
\node[box, below=9mm of s3] (s4) {4\\test};
\node[box, left=of s4] (s5) {5\\elicit};
\node[box, left=of s5] (s6) {6\\settle};
\node[box, left=of s6] (s7) {7\\verdict, score};
\node[rec, left=of s7] (out) {TraceRecord\\written};
\draw[arr] (in) -- (g);
\draw[arr, draw=green!45!black]
  (g) -- node[above, font=\scriptsize, text=green!40!black]{argument} (s1);
\draw[arr] (s1) -- (s2);
\draw[arr] (s2) -- (s3);
\draw[arr] (s3) -- (s4);
\draw[arr] (s4) -- (s5);
\draw[arr] (s5) -- (s6);
\draw[arr] (s6) -- (s7);
\draw[arr] (s7) -- (out);
\end{tikzpicture}
\caption{The reference writer. The formulation gate admits text that
formulates an argument and rejects the rest at negligible cost; stages 1
to 7 construct, test, settle, and score the accepted argument's TRACE
record. Stage 5 is gated by flags, stakes, and budget; contracts,
algorithms, and cost are in Appendix~\ref{app:writer}.}
\label{fig:ocr-pipeline}
\end{figure}

This section depicts the reference writer's stages in operation: the
formulation gate, one claim traced from sentence to verdict, the gated
debate of stage 5, and the rules that stop an evaluation. Figure~\ref{fig:ocr-pipeline} shows the writer whole: the
formulation gate decides whether the text formulates an argument at all,
and stages 1 to 7 then construct, test, settle, and score the record. The
stage table, the two algorithms, the detail tables, and the cost model are
in Appendix~\ref{app:writer}.

\subsection[The Formulation Gate]{The Formulation Gate: Detecting and Typing Reasoning in Narrative}
\label{sec:ocr-detection}

The pipeline of this paper begins with a typed claim. But narratives do not
arrive typed; most text is narration, description, or reporting, and only some
segments \emph{conduct reasoning}. The evaluator therefore needs a stage-zero
operator,
\[
\mathcal{R}: \text{text} \longrightarrow
\{(\text{segment}, \text{claim}, \text{premises}, \text{markers},
\text{family}, \text{scheme}, \text{confidence})\},
\]
which segments the narrative, decides which segments reason, and routes each
reasoning segment to its family-specific tests. Detection rests on three
signal classes, in decreasing order of reliability.

\begin{enumerate}[leftmargin=1.7em]
  \item \textbf{Structural dependence (the criterion).} A segment conducts
  reasoning if and only if it advances a claim whose acceptance is presented
  as depending on other assertions in the segment; that is, Toulmin roles
  (claim, grounds, warrant) are assignable~\cite{toulmin1958uses}. Narration
  and description fail this test: they assert without staking acceptance on
  support.
  \item \textbf{Inferential markers (the flag).} \emph{Therefore},
  \emph{because}, \emph{it follows}, \emph{suggests that}, \emph{must have},
  \emph{if\dots then}, and concessives (\emph{although\dots nevertheless}).
  Markers are sufficient to flag but not necessary: enthymemes carry no
  markers, which is why the charity delta
  belongs partly to detection: the
  reconstruction cost of an unmarked argument is itself a detection output.
  \item \textbf{Family signatures (the router).} Each reasoning family has a
  lexical--modal profile that both identifies it and routes it; the surface
  signature column of Table~\ref{tab:ocr-reasoning-families} lists these
  profiles.
\end{enumerate}

Two properties make this stage more than preprocessing. First, it is
\emph{measurable}: segmentation and typing can be scored against annotated
argument corpora with standard extraction
metrics~\cite{lippi2016argumentation,lawrence2019argument}, giving the
success-criteria table of Section~\ref{sec:ocr-measurement} a row it
otherwise lacks: claim-type accuracy presupposes detection accuracy. Second,
the causal rung markers of Table~\ref{tab:ocr-reasoning-families} make
\textbf{rung collapse detectable at the surface}: a segment whose premises carry rung-1 signatures and whose
conclusion carries rung-2 modality (``is associated with\dots therefore we
should administer'') is flagged before any graph is drawn. This cheap test
catches a large share of causal overclaims at intake, reserving the full
identification machinery of Section~\ref{sec:ocr-pearl-map} for claims that
survive it.

\subsection[From Sentence to Verdict]{From Sentence to Verdict: One Claim Through the Whole Pipeline}
\label{sec:ocr-endtoend}
 
The preceding sections introduced many instruments: a detection operator, a
claim anatomy, reasoning families, argument schemes, causal rungs,
debate-ness conditions, and convergence metrics. A reader meeting them for
the first time can reasonably ask how they fit together on an actual
sentence. This section answers by tracing one claim end to end, the way a
compiler text traces one small program from source code to machine code, so
that every operator is seen doing its one job on the same artifact. The input is deliberately ordinary:
 
\begin{quote}
\emph{``Students who take music lessons score higher on math tests, so
schools should fund music programs to raise math scores.''}
\end{quote}
 
The sentence is a good specimen precisely because it is so agreeable. It
compresses an association, a causal leap, and a policy demand into one
fluent breath. As we will see at Stage~5, its conclusion sits squarely
in the modal basin: it is what an unconditioned language model is most
likely to say. The pipeline's job is to take it apart.
 
\paragraph{The formulation gate
(Section~\ref{sec:ocr-detection}).} The gate admits the segment and, in
admitting it, already finds the story: the premise carries a rung-1
signature (\emph{score higher}, an observed association) while the
conclusion carries rung-2 modality (\emph{fund\dots to raise}) wrapped in
practical-deontic \emph{should}, so the surface rung-collapse flag fires
before any deeper machinery runs, and the segment is routed to three
batteries at once: inductive, causal, and practical.
 
\paragraph{Stage 1: CRIT intake and Toulmin anatomy
(Sections~\ref{sec:ocr-reasoning-vs-debate}, \ref{sec:ocr-argtheory}).}
CRIT fills the TRACE record. Claim \(\Omega\): schools should fund music
programs \emph{in order to raise math scores}. The italicized purpose
clause matters, because it makes the policy conditional on a causal
mechanism. Grounds: the observed score gap between students who do and do
not take lessons. Warrant (hidden, now surfaced): \emph{the association
exists because music training causes mathematical improvement, and what
causes improvement should be funded}. Backing: none offered. Qualifier
mismatch: the grounds are a population-level tendency; the conclusion is an
unqualified prescription. Rebuttal: absent; the sentence considers no
alternative explanation and no downside. Already the anatomy shows the
argument is an enthymeme with a critical hidden warrant; the charity
delta will be large.
 
\paragraph{Stage 2: Family typing
(Section~\ref{sec:ocr-reasoning-families}).} The single sentence
decomposes into three typed subclaims, each owed a different standard:
\(S_1\) (inductive/statistical): music students score higher, a claim
checkable against data; standard: sample adequacy, reference class. \(S_2\)
(causal, rung 2): funding music lessons would \emph{raise} scores; standard:
identification. \(S_3\) (practical/normative): therefore schools should
fund them; standard: means--end fit, alternatives, side effects, plus a
value premise about how school resources ought to be allocated. Collapsing
the three into one verdict is precisely the error the paper is built to
prevent: \(S_1\) may be true, \(S_2\) unidentified, and \(S_3\) unjustified,
all at once.
 
\paragraph{Stage 3: Scheme and critical questions
(Section~\ref{sec:ocr-argtheory}).} The argument instantiates two Walton
schemes in series: \emph{correlation to cause} (from \(S_1\) to \(S_2\)) and
\emph{practical reasoning from positive consequences} (from \(S_2\) to
\(S_3\)). The critical questions arrive pre-written. For correlation to
cause: could a third factor produce both variables? (Family income and
parental involvement predict both music lessons and math scores.) Could
causation run in reverse? (Mathematically stronger students may be steered
toward music.) Would the relation survive intervention? (Unknown; that is
exactly \(S_2\).) For practical reasoning: are there alternative actions
that achieve the goal at lower cost? (Direct math tutoring is an obvious
rival.) What are the side effects and what is displaced by the funding? The
scheme layer has converted a vague unease into an explicit, finite question
list, and the record now shows which questions the original argument
answered: none.
 
\paragraph{Stage 4: Pearl typing and identification
(Section~\ref{sec:ocr-pearl-map}).} The causal subclaim \(S_2\) is typed
formally: the grounds report \(P(\text{math} \mid \text{music})\); the
conclusion needs \(P(\text{math} \mid do(\text{music}))\). With plausible
confounders (household income, parental involvement, school quality)
unmeasured in the ledger, the interventional quantity is not identified
from the observational association; the honest output is a bounded
statement, not a confident
effect~\cite{manski1990nonparametric,balke1997bounds}. The evaluator names
what would identify it: randomized assignment of music instruction, a
lottery-based natural experiment, or within-family comparisons. It also
records that controlled studies of music training have generally failed to
find the far-transfer effect on academic outcomes that the argument
assumes~\cite{sala2020music}. Note what just happened: the pipeline did not
declare the claim false; it declared it \emph{unwarranted at the requested
rung given the present ledger}, which is a different and more useful thing.
 
\paragraph{Stage 5: The basin, and the debate that escapes it.} Why is a procedure needed at all? Would a
single model not notice all this if simply asked? Often, no, and the
maximum-likelihood trap explains why. \emph{Music makes children better at
math} is a culturally beloved claim; text asserting it vastly outnumbers
text dissecting it, so the mode of
\(P_\theta(\cdot \mid \text{sentence})\) is fluent elaboration, the
Mozart-effect genre, rather than critique. A single unconditioned pass
tends to stay in that basin, and \(k\) unconditioned passes agree with one
another because they sample from near-identical distributions, a point
Section~\ref{sec:ocr-debate-layer} makes precise: agreement among strongly
correlated belief generators is cheap. Role conditioning changes the distribution being
sampled. A proponent, a skeptic instructed that its success criterion is
finding faults, and a moderator holding contentiousness high force
low-probability, coherent continuations into the exchange: the confounder
objection, the reverse-causation objection, the tutoring alternative, the
null experimental literature. EVINCE's dials show the signature of a
healthy episode: cross-agent divergence \emph{rises} while the hypothesis
set grows, then falls as audited evidence units, not concessions, enter
the ledger; the warranted convergence ratio, a metric
defined in Appendix~\ref{app:convergence}, stays high because each belief
movement co-occurs with a new ledger entry. The debate need not create
new knowledge; often it moves probability mass toward stored or retrievable
knowledge that an unconditioned pass would not surface.
 
\paragraph{Stage 6: Attack graph and acceptability
(Section~\ref{sec:ocr-argtheory}).} The exchange leaves a small argument
graph. \(A\): the original argument. \(B\): the confounding objection,
attacking \(A\)'s warrant. \(C\): the reverse-causation objection, also
attacking the warrant. \(D\): the experimental-evidence argument (null
far-transfer results), attacking \(S_2\) directly. \(E\): a reply to \(B\)
(``some studies control for income''), attacking \(B\); but CRIT's audit
finds \(E\)'s grounds weak (the controls are partial), so \(E\) does not
defeat \(B\). Under grounded semantics, \(B\), \(C\), and \(D\) are
acceptable; \(A\) is not: every path to reinstating \(A\) requires
defeating all three attackers, and none is defeated. The verdict is
computed from the graph and the audited ledger, not voted by the
debaters; that is external adjudication (D4) in action.
 
\paragraph{Stage 7: Verdict, repair, and the metric readout
(Sections~\ref{sec:ocr-stopping}--\ref{sec:ocr-measurement}).} The typed
verdict separates the three subclaims. \(S_1\): provisionally accepted if the
reported association is supported by the supplied data. \(S_2\): defer, underidentified from the
ledger; the named missing evidence is randomized or quasi-experimental
assignment; the prior experimental record points against far transfer.
\(S_3\): rejected as stated, because a practical conclusion cannot stand on an
unidentified causal premise, and its critical questions (alternatives, side
effects, value premise) went unanswered. The repaired argument a proponent
could honestly make is much more modest: \emph{music programs may be worth
funding for their intrinsic and demonstrated musical benefits, but not on
the promise of raising math scores; if raising math scores is the goal,
directly targeted interventions have stronger evidence}. The distance
between the original and this repair is the charity delta, and it is
recorded. The episode also emits its convergence diagnostics: the stop
occurred when the log-odds on \(S_2\) crossed the defer threshold
(Proposition~\ref{prop:ocr-sprt}), procedural-invariance reruns with
permuted roles reproduce the verdict, and the ledger alone, without the
transcript, suffices for a cold judge to re-derive it.
 
Readers who want the same pipeline at clinical stakes will find the
illustration in Appendix~\ref{app:illustrations}.

\paragraph{What the trace shows.} The pipeline's value in this trace did
not come from creating new knowledge. It came from redistributing
probability mass so that stored, retrievable, or otherwise available
knowledge that an unconditioned pass would not surface gets said, audited,
and counted. That is the paper's argument in miniature: reasoning quality
is not a property of fluent output; it is a property of a record that
survives inspection. The trace is Algorithms~\ref{alg:ocr-trace}
and~\ref{alg:ocr-trace2} executing line by line; the record states above
are their intermediate values.

\paragraph{One architecture, interchangeable plug-ins.} The trace ran the
causal plug-in because Pearl's hierarchy gives causal reasoning the most
mature formal semantics, but the architecture is family-neutral: every
family passes the same gate and stages and contributes its own test battery
and outputs. The causal instance is representative, not privileged. The
plug-in inventory and the causal module mapping are
Tables~\ref{tab:ocr-plugins} and~\ref{tab:ocr-pipeline} in Appendix~\ref{app:writer}.

\subsection{Stage 5: Gated Debate}
\label{sec:ocr-debate-layer}

Stage 5 is the writer's only unbounded-cost stage, and it is therefore
gated: debate runs only when a hard flag, the stakes, and the budget
jointly warrant it. When it runs, its purpose is elicitation under
opposition, surfacing the evidence, alternatives, and defeaters that an
unconditioned pass leaves inside the modal basin. Exchange carries
evaluative force only when four properties hold:
\label{sec:ocr-debateness}

\begin{enumerate}[leftmargin=1.7em]
  \item[\textbf{D1}] \textbf{Role separation.} Proponent and critic occupy
  distinct contexts, not merely distinct sentences in one context window.
  \item[\textbf{D2}] \textbf{Objective opposition.} The critic's success
  criterion is finding faults, not reaching agreement. Sycophancy induced by
  preference training attacks D2 directly~\cite{sharma2024sycophancy}.
  \item[\textbf{D3}] \textbf{Epistemic diversity.} The parties' priors and
  likelihoods are not identical; otherwise agreement is preordained and
  uninformative.
  \item[\textbf{D4}] \textbf{External adjudication.} A standard outside the
  disputants (a judge, an evidence ledger, a formal checker, a CRIT audit)
  settles the verdict; the disputants' own consensus does not.
\end{enumerate}

These four properties define a debate-ness profile on which every
reasoning-as-debate method sits, and the empirical record orders itself
along it (Table~\ref{tab:ocr-debateness}, Appendix~\ref{app:convergence}): methods that realize none of
the four fail to self-correct, same-model panels that realize only D1
degenerate toward premature agreement, and protocols that enforce D2 and D4
earn formal guarantees. The SocraSynth, EVINCE, and CRIT line of this
research program is an engineering program for restoring D2, D3, and D4 to exchanges
that would otherwise possess only
D1~\cite{chang2024socrasynth,chang2024evince,chang2023crit}.

Classical agreement results issue the right warning first. Agents with a
common prior cannot agree to disagree~\cite{aumann1976agreeing}; iterated
communication reaches consensus that need not aggregate all private
information~\cite{geanakoplos1982disagree}; and mechanical opinion averaging
converges with no truth content at all~\cite{degroot1974consensus}.
Multi-LLM panels, whether instantiated from one base model or from several
vendors trained on overlapping corpora, are strongly correlated belief
generators, so convergence among them should be treated as carrying little
evidential weight unless it is coupled to new audited evidence. Agreement is
cheap. The consequence shapes every metric below: \textbf{the object that
carries evidential weight is not the consensus but the ledger}, the set of
typed, CRIT-audited evidence units surfaced during the exchange, together
with the trajectory by which beliefs moved. Every metric that follows is a
ledger or trajectory metric, not an agreement metric.
The validity metrics that make these properties measurable, procedural
invariance, warranted convergence, evidence-counterfactual sensitivity, the
dual-phase signature, effective panel size, and the external-judge gap, are
defined in Appendix~\ref{app:convergence}, together with the imported
agreement and optimality results; the companion volume computes them.

One boundary closes the stage. Debate is a search procedure: it can expose
assumptions, improve evidence coverage, and reduce overclaiming, but it
cannot by itself certify truth, and a converged verdict remains a proposal
that must pass the admission control appropriate to its domain before any
durable state changes. The pipeline improves proposals; it does not make
them self-authorizing.

\subsection{Stopping Conditions}
\label{sec:ocr-stopping}

A reasoning controller should not run until agents merely agree. The four conditions below are not ad hoc: they are the qualitative face of a sequential test. Treating each turn's audited evidence as an increment to the log-odds of the claim, the controller stops when the accumulated log-odds cross thresholds set by the costs of the two errors, thresholds that ERM's regret asymmetries supply directly (Proposition~\ref{prop:ocr-sprt}). What follows is the operational statement of that rule; the controller stops when one of the following conditions holds:

\begin{proposition}[Optimal stopping]
\label{prop:ocr-sprt}
Under the idealization that turns yield conditionally independent evidence
increments for a binary claim, the sequential probability ratio test with
thresholds \(A=\log\frac{1-\beta}{\alpha}\) and
\(B=\log\frac{\beta}{1-\alpha}\), with error rates \(\alpha,\beta\) chosen
from the regret asymmetries priced by ERM, minimizes the expected number of
turns among all stopping rules with equal error probabilities
\cite{wald1945sequential,wald1948optimum}. The four stopping conditions of
Section~\ref{sec:ocr-stopping} are operational counterparts of this
sequential-testing view: qualified proposal and rejection correspond to
threshold crossings, diminishing returns to vanishing increments, and defer
to a value-of-information judgment that further exchange is not worth its
cost.
\end{proposition}

\begin{enumerate}[leftmargin=1.7em]
    \item \textbf{Qualified proposal:} the claim is typed, grounded, audited, calibrated, and has no unresolved high-severity objection.
    \item \textbf{Rejected claim:} the claim fails claim typing, scheme fit, premise support, source credibility, causal rung discipline, identifiability, or contradiction checks.
    \item \textbf{Defer:} the evaluation exposes missing evidence, underidentification, unresolved defeaters, or value-context ambiguity that cannot be resolved with available information.
    \item \textbf{Diminishing returns:} additional exchange produces low evidence gain or repeated arguments without new information.
\end{enumerate}

The defer outcome is essential. A reasoning system must be able to say that the evidence is insufficient, the causal quantity is underidentified, the legal standard is unclear, the value frame is unresolved, or the intervention is not defined well enough to support action. A useful defer decision reports more than \emph{unknown}: it reports the unresolved claim type, missing variables or assumptions, relevant bounds if available, and the next evidence that would most reduce uncertainty.

\section{Measuring the Writer and the Record}
\label{sec:ocr-measurement}

The pipeline is evaluable because each stage produces observable outputs. The metrics below (Table~\ref{tab:ocr-reasoning-success}) are not measurements of objective reasoning quality; they measure record completeness, writer behavior, gate behavior, and consumer usefulness. The objective is not agreement, but improvement of the TRACE record.

\begin{table}[H]
\centering
\small
\begin{tabular}{p{0.28\linewidth}p{0.62\linewidth}}
\toprule
\textbf{Dimension} & \textbf{Success criterion} \\
\midrule
Claim-type accuracy & The evaluator correctly classifies the claim as factual, causal, practical, legal, diagnostic, normative, or mixed. \\
Scheme fit & The argument scheme is correctly identified and matched to the right critical questions. \\
Premise and evidence support & Premises are evidenced at the standard required by the claim type and context. \\
Source credibility & Testimony, expertise, quotations, and external claims are traced and credibility-rated. \\
Rung discipline & Embedded causal claims correctly distinguish association, intervention, and counterfactuals. \\
Alternative recall & Plausible competing explanations and counterarguments are surfaced. \\
Overclaim control & The conclusion does not exceed the premises, evidence, causal rung, or value frame. \\
Calibration & Confidence after evaluation better matches evidential support. \\
Regret reduction & The expected cost of remaining reasoning error is reduced. \\
Recurrence control & The system makes fewer similar reasoning mistakes in later episodes. \\
\bottomrule
\end{tabular}
\caption{Writer and record-quality criteria. Debate can help improve these dimensions, but the dimensions evaluate the record and its writer, not debate performance.}
\label{tab:ocr-reasoning-success}
\end{table}

Success criteria become measurements only when each dimension has a unit, a
label source, a scoring rule, and a declared role in the verdict.
Table~\ref{tab:ocr-metric-defs} (Appendix~\ref{app:bench}) supplies these operational definitions; the
gate column marks the dimensions that participate in the hard gates of the
final status rather than in the optimized residual.

Four boundary notes on the gate column. First, DetectionF1 is marked as a
precondition rather than a gate: TypeAcc is meaningful only over the
segments detection found, so a weak detector invalidates the rest of the
vector before any gate applies. Second, claim typing gates operationally
even though TypeAcc does not gate benchmarks: in high-stakes settings, when
type confidence falls below a declared threshold the correct action is to
defer or route to an external adjudicator, because a mistyped claim makes
every downstream verdict malformed. Third, signed quantities such as
CalibrationGain and RegretReduction are normalized before aggregation;
negative values are reported explicitly and clipped or rescaled according to
the task's declared scoring convention. Fourth, defects are counted once:
when a critical question is formally subsumed by a family test, as the
correlation-to-cause questions are by the causal battery, the failure is
charged to the family metric (here RungCollapseRate or BoundViolationRate),
and CQCoverage records the question as answered by the plug-in rather than
charging SchemeFit a second time.

Several metrics can be estimated with existing or constructible benchmarks. Rung accuracy and rung-collapse rates can be measured on tasks organized by association, intervention, and counterfactual reasoning; CausalT5K, CLadder, and correlation-to-causation benchmarks provide examples of this style of evaluation~\cite{causalT5K,jin2023cladder,jin2023corr2cause}. Calibration improvement can be measured with proper scoring rules, Brier score, or expected calibration error~\cite{guo2017calibration}. Clinical examples require expert annotation, but the same structure applies: expert labels define relevant alternatives, missing tests, overclaims, and high-regret errors.

The benchmark and ablation protocol that operationalizes these metrics and
gates, TRACE-Bench, is specified in Appendix~\ref{app:bench}; its
baselines include the single-pass
writer as the staged decomposition's null hypothesis and ungated debate,
this research program's earlier evaluation mechanism, as the within-family
comparison.

\section{Claims, Non-Claims, and Falsification}
\label{sec:ocr-falsify}

The paper's claims should be stated at their exact strength, because both
overclaiming and underclaiming misdescribe the design.

What is claimed. TRACE is not defended as an optimal eight-stage reasoning
evaluator or as a universal reasoning-quality metric. It is defended as a
typed, versioned reasoning-log schema and admission-outcome explanation
contract for agentic systems: the schema and the stage contracts are the
contribution, the pipeline is one implementation for writing the log, and
the value is measured by whether downstream consumers, Trivium,
MACI, ALAS, SagaLLM \cite{chang2025sagallm}, and Mnemosyne \cite{chang2027mnemosyne}  make better, more auditable, more repairable
decisions when they consume the records. TRACE still performs
evaluation-like functions; what is disclaimed is only the pretension that
its verdicts measure reasoning quality against an objective standard. The
scores measure compliance with the declared, versioned TRACE policy, not
objective reasoning quality. Within that policy, some conditional verdicts are largely
model-independent once their representation is declared: identification
against a declared graph is decided by do-calculus, deductive validity by a
proof or entailment check, and acceptability by a deterministic semantics
over a declared attack graph. The representation boundary, the graph,
subclaims, warrants, and attack edges themselves, remains model- or
human-mediated and is therefore measured separately by typing, extraction,
and human-audit agreement. In addition, the perturbation metrics of
Section~\ref{sec:ocr-validity-metrics}, procedural invariance and
evidence-counterfactual sensitivity, require neither gold labels nor policy
weights. Policy dependence is confined to the aggregate scalar of
Equation~\eqref{eq:ocr-reasoning-quality}, which is a reporting device and
is never reported without its vector and failed gates.

\paragraph{The consumer claims.}
\label{sec:ocr-consumers}
Two consumer-facing claims complete the set, and both are central to
the falsification criteria below. Across all four consumers, the constant contribution is not that the accept
or reject decision beats a baseline gate; on many items it may tie. It is
that every outcome, admit, reject, and defer alike, arrives with a typed
account of why, on what evidence, and what would change it. This claim is
uniform, it survives wherever gating accuracy ties, and it is operational in
three measurable senses. Faithful: the verdict is reconstructable from the
record alone by a cold judge, and it moves under evidence ablation, so the
stated reasons are decisive rather than decorative. Actionable: repairs,
when applied, flip verdicts at a measurable rate, and defers name evidence
that resolves them when supplied. Consequential: systems that learn from
records outperform systems that learn from scalar outcomes on downstream
task error and error recurrence. The closed-loop experiment of the implementation
plan, two identical agents on a multi-week task stream, one writing
memory through the gate and one writing freely, measures the third sense
directly, and it is the headline test of the schema's consumer value.

Here a division of labor
must be stated to avoid an overclaim. Structural admission, whether the
workflow is well formed, its dependencies acyclic, its resource and temporal
constraints stated and satisfiable, its preconditions met, belongs to the
plan validator and needs no reasoning evaluator; much of it is checkable by
solvers and consistency checks. TRACE gates the justificatory residue: the
claims a plan rests on that constraint checking cannot reach. Why this
branch rather than that one; whether the causal premise behind an action
supports the expected effect at the rung the plan requires; whether an
assumed precondition is evidence-backed or merely asserted. TRACE audits the
plan's warrants, not its wiring. The measured quantity is accordingly the
prevented-error rate on plans that pass structural validation yet fail
justification, together with the added latency priced against the cost model
of Section~\ref{sec:ocr-trace-method}.

\subsection{Illustrative Consumer Vignette: Flood Search-and-Rescue}
\label{sec:ocr-flood-vignette}

\begin{figure}[htbp]
\centering
\includegraphics[width=0.86\linewidth]{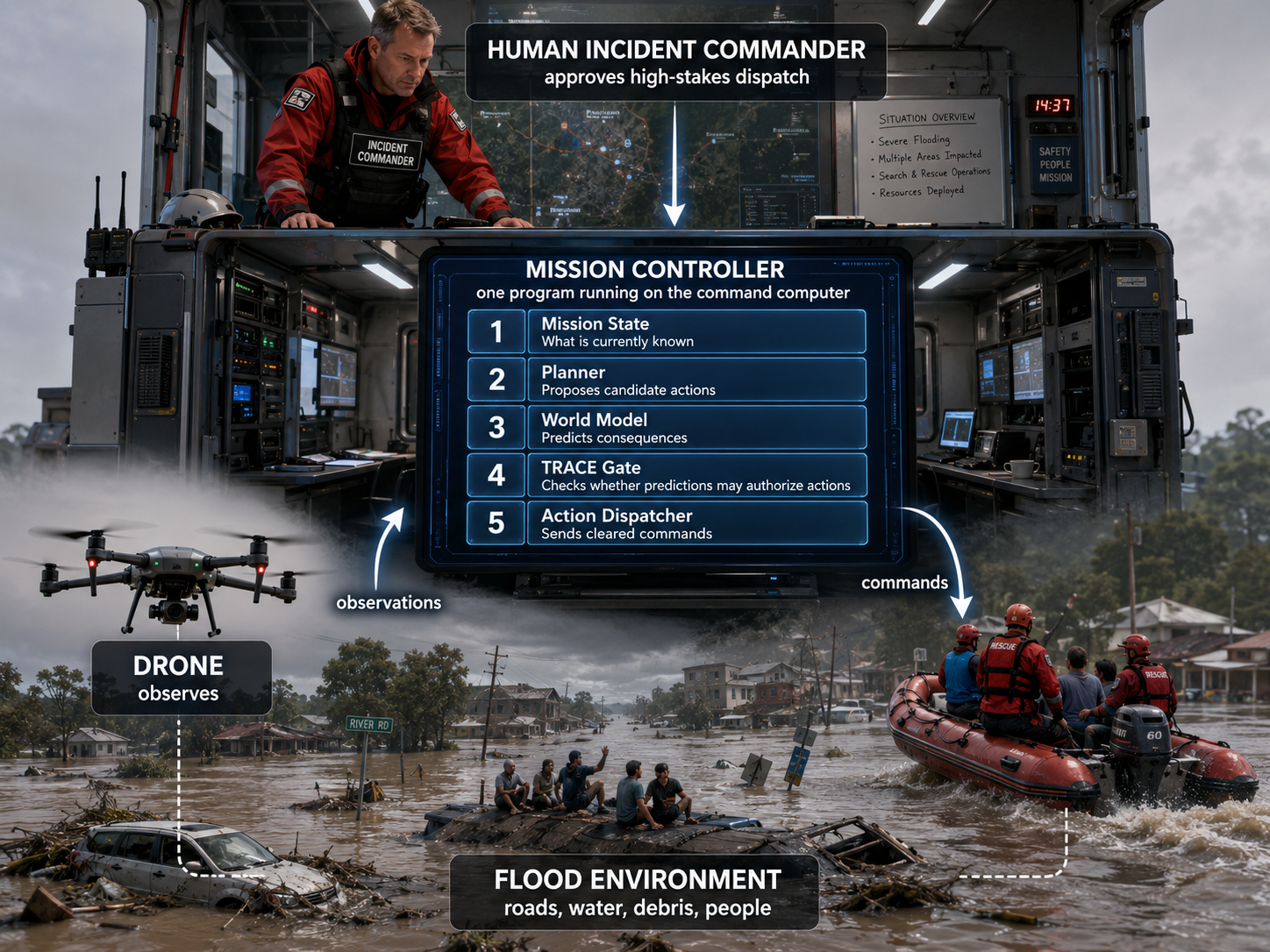}
\caption{The flood search-and-rescue setting for the vignette. A single
Mission Controller runs the mission state, planner, predictive world model,
TRACE gate, and action dispatcher on the command computer; a survey drone
supplies observations; a rescue boat executes an authorized dispatch; and a
human Incident Commander approves high-stakes actions. TRACE neither flies the
drone nor drives the boat: it audits, at the gate, whether a prediction may
authorize an irreversible dispatch. The scene is illustrative and the
consequences are simulated. The scene is an AI-generated illustration.}
\label{fig:ocr-flood-scene}
\end{figure}

The music-lessons trace shows how the writer types and evaluates an argument.
A second illustration is needed to show why the resulting record is useful to
a consumer. Consider a simulated flood-response mission (Figure~\ref{fig:ocr-flood-scene})
with one central
\emph{Mission Controller}, one survey drone, one rescue boat, and one human
Incident Commander. The Mission Controller contains the mission state,
planner, predictive world model, TRACE gate, and action dispatcher. The drone
observes; the boat executes an authorized rescue; and the Flood Environment
supplies observations and consequences. Simulation ground truth may contain
facts that the Mission Controller has not observed. In particular, the
environment contains a debris obstruction in the North Channel, while the
Mission Controller legitimately knows only that the route is unverified.
Hidden simulation state is not a premise available to the writer.

Four residents are waiting at Riverside Apartments. The planner proposes
three structurally valid candidates: dispatch the boat through the short North
Channel, send the drone to inspect that channel, or dispatch the boat through
a longer observed-open South Detour. A prediction service reports a plan
success value of $0.92$ for the northern dispatch. The same evidence object
records model support of $0.28$ and an out-of-distribution score of $0.82$.
These values are deterministic fixtures chosen to expose the record logic;
they are not experimental measurements. The distinction they make concrete is
decisive: a high reported success value does not license action when the
forecast was produced outside its declared support. Two record mechanisms act
together on the northern claim, and the vignette keeps them distinct: a failed
hard support gate \emph{forbids} clearance now, while the \texttt{defer}
status with a named \texttt{missing} field and a bounded \texttt{repair}
makes the hold \emph{resolvable} rather than terminal. A support failure is
thus not a bare refusal; it is a refusal that carries the evidence request
that could lift it.

Table~\ref{tab:ocr-flood-records} follows the records rather than the agents.
The predictive route claim, the bounded evidence-seeking action, the revised
route claim, and the supported alternative each receive their own TRACE
record because they are different claims with different warrants. The
consumer decisions are not TRACE verdicts: \textsc{Hold} and \textsc{Clear}
are plan-gate actions written to \texttt{consumer\_actions} after the
consumer reads the record.

\begin{table}[H]
\centering
\scriptsize
\begin{tabular}{@{}p{0.09\linewidth}p{0.24\linewidth}p{0.30\linewidth}p{0.27\linewidth}@{}}
\toprule
\textbf{Record} & \textbf{Typed claim and TRACE status} & \textbf{Decisive record fields} & \textbf{Consumer consequence} \\
\midrule
$R_N^{(1)}$ & Predictive: North Channel is traversable and the boat will arrive before the deadline. \texttt{defer} on a failed support gate. & Reported success $0.92$; support $0.28$; OOD $0.82$; failed hard support gate (OOD above threshold); \texttt{missing}: current route observation; \texttt{repair}: bounded drone verification. & Hold the irreversible northern dispatch. The failed support gate forbids clearance now; the defer is actionable because it names the observation that could move the claim back inside support. \\
\addlinespace[2pt]
$R_V^{(1)}$ & Practical: the drone can inspect North Channel and return within its admitted envelope. \texttt{accept} or \texttt{qualify}. & Battery, weather, reversible scope, observation target, and any authority requirement. & Clear only the bounded verification action; do not treat it as indirect clearance of the boat dispatch. \\
\addlinespace[2pt]
$R_N^{(2)}$ & Predictive revision: North Channel is traversable. \texttt{reject}. & New drone observation of the obstruction; contradiction flag; \texttt{revision\_history} link to $R_N^{(1)}$. & Block the northern dispatch while preserving the original forecast and the reason it was held. \\
\addlinespace[2pt]
$R_S^{(1)}$ & Predictive and practical: South Detour is traversable and the boat can complete the pickup. \texttt{accept} or \texttt{qualify}. & Current route report, capacity, deadline, policy version, and external Incident Commander approval where required. & Clear the structured southern dispatch and cite the exact authorizing record version. \\
\bottomrule
\end{tabular}
\caption{An illustrative TRACE history for flood search-and-rescue. The
numbers are pedagogical fixtures, not results. The example demonstrates how
one schema represents defer, evidence-seeking repair, append-only revision,
and a later consumer action without treating prediction as permission.}
\label{tab:ocr-flood-records}
\end{table}

\begin{figure}[H]
\centering
\begin{tikzpicture}[
  rec/.style={draw=violet!60!black, fill=violet!8, rounded corners=3pt,
              align=center, font=\scriptsize, text width=26mm, inner sep=3pt},
  act/.style={draw=blue!55!black, fill=blue!7, rounded corners=3pt,
              align=center, font=\scriptsize, text width=22mm, inner sep=3pt},
  obs/.style={draw=green!45!black, fill=green!10, rounded corners=3pt,
              align=center, font=\scriptsize, text width=22mm, inner sep=3pt},
  arr/.style={->, >=stealth, thick, draw=gray!55!black},
  node distance=6mm and 6mm
]
\node[rec] (rn1) {$R_N^{(1)}$\\predictive claim\\\texttt{defer}\\missing: route view};
\node[act, right=of rn1] (hold) {plan consumer\\\textsc{Hold} boat};
\node[rec, below=9mm of rn1] (rv1) {$R_V^{(1)}$\\practical claim\\verification\\\texttt{accept/qualify}};
\node[act, right=of rv1] (drone) {dispatch\\survey drone};
\node[obs, right=of drone] (obs) {new observation\\North Channel blocked};
\node[rec, right=of obs] (rn2) {$R_N^{(2)}$\\revision\\\texttt{reject}\\preserve $R_N^{(1)}$};
\node[rec, below=9mm of rn2] (rs1) {$R_S^{(1)}$\\South Detour claim\\\texttt{accept/qualify}};
\node[act, left=of rs1] (clear) {human approval, then\\\textsc{Clear} boat};

\draw[arr] (rn1) -- (hold);
\draw[arr] (rn1) -- node[left, font=\scriptsize]{repair} (rv1);
\draw[arr] (rv1) -- (drone);
\draw[arr] (drone) -- (obs);
\draw[arr] (obs) -- (rn2);
\draw[arr] (rn2) -- node[right, font=\scriptsize]{local replan} (rs1);
\draw[arr] (rs1) -- (clear);
\end{tikzpicture}
\caption{From prediction to attributable action. A defer names an
information-gathering repair; the repair has its own record; new evidence
creates an append-only revision; and a later record licenses a different
route. TRACE neither supplies the planner nor controls the drone or boat. It
records the justificatory boundary through which their outputs become durable
commitments.}
\label{fig:ocr-flood-lifecycle}
\end{figure}
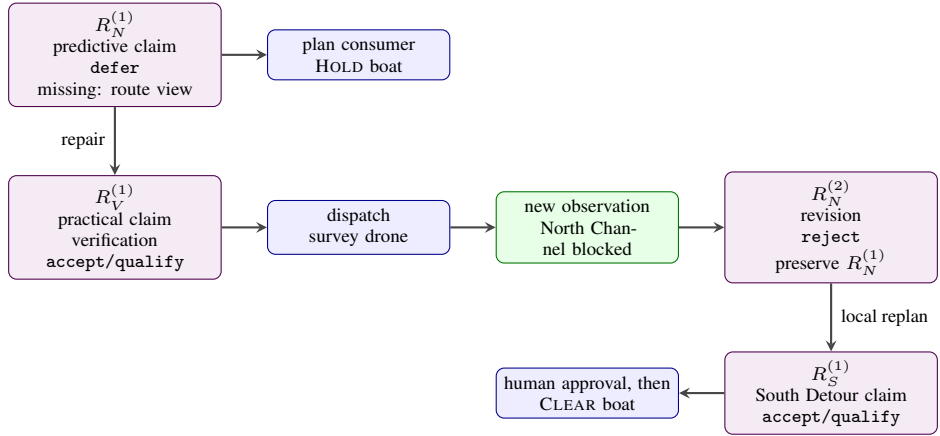

The vignette complements the music-lessons trace of
Section~\ref{sec:ocr-endtoend}. The music example runs an argument through the
writer and shows why association, intervention, and practical prescription
must be typed separately. The flood example starts after candidate plans
exist and shows why the resulting record needs \texttt{failed\_gates},
\texttt{missing}, \texttt{repair}, \texttt{revision\_history}, and
\texttt{consumer\_actions}. It also makes the structural--justificatory
division concrete: the planner and validator decide whether a route and asset
assignment are well formed; TRACE audits whether the claims licensing that
commitment are sufficiently warranted. Appendix~\ref{app:maci} gives the
same vignette as a plan-gating usage scenario.

This example is a specification illustration, not evidence that TRACE
improves rescue performance or that any predictive model is accurate. A
future closed-loop study must hold the flood environment, observations,
candidate plans, predictive outputs, policies, and scenario seeds fixed while
varying the audit substrate. Pre-registered outcomes should include
unsupported-commitment rate, false-hold rate, decision latency, record
reconstructability, repair success, and localization of invalidated plan
branches.

The vignette also clarifies how the paper's measurements should be read,
because the same record fields it exercises, verdicts, gates, evidence,
warrants, carry very different evidential weight. The metric vector should
accordingly be read in epistemic tiers, from
strongest to weakest warrant. Tier one is formally checkable: rung
discipline, identification status, bound violations, and deductive validity
have ground truth that is a theorem or an oracle model, not an annotator's
opinion. Tier two is checkable without any ground truth: the perturbation
metrics ask whether a verdict survives role, order, and seed permutation and
whether it moves when decisive evidence is ablated and stays when only
rhetoric is paraphrased; a memorized or sycophantic verdict is unlikely to
survive a sufficiently rich perturbation suite, because the expected verdict
transitions are specified at the level of decisive evidence, warrants,
and gates rather than surface labels, which makes this tier the
contamination-resistant core of TRACE-Bench, resistant by test design rather
than by secrecy. Tier three is
moderate-agreement annotation: detection, anatomy, and scheme fit, bounded
by inter-annotator agreement and reported with it. Tier four is the
contested holistic scalar, which inherits the disagreement of its annotators
and is quarantined by the label-source declarations of
Table~\ref{tab:ocr-metric-defs} (Appendix~\ref{app:bench}). The paper's warrant lives in tiers one
and two.

What is not claimed. The following are deliberately left open, and stating
them as open is part of the contract:

\begin{enumerate}
    \item The eight-stage decomposition is not claimed to be optimal;
    stages may merge or split as the ablations direct.
    \item The family taxonomy is not claimed to be complete.
    \item The score vector is not claimed to be uniquely correct, and its
    weights are not claimed to be universal; they are governance decisions,
    versioned and owned by the governance layer (DIKE and ERIS) of the
    companion volume.
    \item The model-mediated stages are not claimed to execute reliably;
    their reliability is measured, not assumed.
    \item The convergence metrics are not claimed to correlate with truth;
    they measure stability and responsiveness.
    \item The benchmark is not claimed to represent all reasoning.
\end{enumerate}

Falsification. A decomposed framework has a characteristic failure mode:
every wrong verdict can be blamed on an upstream mis-typing, so the
framework never appears to fail. The defense is to name the judges in
advance. The following criteria are pre-registered as the results that would
count against the design, and no post hoc reattribution to upstream stages
is admissible against them:

\begin{enumerate}
    \item If the single-pass writer matches the staged writer on record
    quality and error attribution at comparable cost, the decomposition is
    unjustified.
    \item If the no-gates ablation shows no separation in WrongAcceptRate
    on fluent but unsupported or formally invalid items, the gates are
    decorative.
    \item If procedural invariance and evidence-counterfactual sensitivity
    fail to separate systems that differ in evaluation discipline, the
    consistency metrics are uninformative.
    \item If evaluator-versus-human agreement at the typing boundary falls
    below its declared threshold, the downstream formal checks are checking
    the wrong objects and the pipeline's warrant collapses with it.
    \item If the record-gated agent of the closed-loop experiment does not
    outperform its ungated twin on downstream task error and error
    recurrence, the record earns no consumer value and the schema is an
    archive, not an instrument.
\end{enumerate}

One baseline in this program deserves its own sentence. Ungated debate
scored holistically is this research program's earlier evaluation mechanism, CRIT
over SocraSynth and EVINCE exchange; it is retained inside stages 5 and 6 as
elicitation and scoring machinery, and retained in TRACE-Bench as a
baseline. The benchmark is therefore also the volume auditing its
predecessor with its own methods, and the comparison is within-family:
what is tested is whether typing and gating improve the instrument that
proposed them.
\subsection{Limits}
\label{sec:ocr-limits}

No TRACE record, writer, or debate system can guarantee complete causal truth or wise action. Several limits are unavoidable:

\begin{enumerate}
    \item Hidden confounders may remain undiscovered.
    \item Evidence may be absent, corrupted, or unobservable.
    \item A causal quantity may be mathematically underidentified.
    \item Counterfactuals may remain unknowable under the available assumptions.
    \item Agents may share priors and therefore share errors.
    \item Correct causal reasoning does not by itself determine wise action.
    \item Gates verify field consistency, not field truth. A deterministic
    gate reads what the record says; a writer that fabricates evidence
    entries or misstates support can populate fields that pass every gate.
    The schema's countermeasures are provenance and source-credibility
    fields that make each entry attributable, and the perturbation suite,
    which tests that verdicts track decisive evidence; but record
    faithfulness is not world truth, and the schema does not claim
    otherwise.
\end{enumerate}

The aim is therefore not certainty. The aim is disciplined improvement:
claims should become more explicit, better grounded, more robust to critique,
less overconfident, easier to revise, and less likely to repeat the same
failure modes over time. The flood vignette adds two further boundaries. Its
simulation truth is not an input to the Mission Controller, and technical
TRACE clearance does not manufacture emergency-command authority. Both are
external constraints on any future implementation.

\section{Summary}
\label{sec:ocr-summary}

The paper's product is an artifact and a discipline. The artifact is the
TRACE record: a typed, versioned schema for logging what was claimed,
what kind of reasoning the claim required, what evidence and warrants were
offered, what was tested, what failed, what is missing, and what verdict
followed, with the TRACE causal record as its specialization for the family
with the most mature semantics. The discipline is closure: no durable state
change without a record.

Around the artifact the paper arranged four things. A type system gives
the fields their semantics: CRIT supplies the Socratic roof; the reasoning
families and the Toulmin, Walton, and Dung toolbox supply claim types,
roles, schemes, and critical questions; Pearl's ladder supplies rung,
identification, and bounds, with association, intervention, and
counterfactuals positioned as one disciplined family inside reasoning rather
than reasoning itself. A reference writer, TRACE, fills the fields under
stage contracts, with cost controlled by a cascade and escalation gated by
flags, stakes, and budget; the writer is one conforming implementation, and
the schema outlives it. A measurement regime holds writer and record to
account: hard gates first, scalar optimization on the residual, the vector
never reported without its failed gates, and TRACE-Bench, specified in
Appendix~\ref{app:bench}, whose baselines include the single-pass writer as the decomposition's null
hypothesis and this research program's earlier evaluation mechanism, ungated debate
scored holistically, as the within-family comparison. And a set of consumers
gives the record its value: memory admission, plan gating on a plan's
warrants rather than its wiring, temporal regret over attributed steps, and
verdict reuse. The music-lessons trace shows the writer separating an
association, an intervention, and a practical conclusion; the flood-rescue
vignette shows a plan consumer using \texttt{missing}, \texttt{repair},
\texttt{revision\_history}, and \texttt{consumer\_actions} to hold one
branch, request evidence, preserve a contradiction, and authorize another.

The final output for any claim remains a qualified proposal, a rejection, or
a defer that names its missing evidence. A fluent consensus can be evidence
that reasoning has improved; it is not truth and it is not authority. What
the record buys is not correctness but attributability: when the system is
wrong, the record says where. Whether that is worth its cost is the
falsifiable question the companion volume's experiments are built to answer.

\appendix

\section{TRACE Schema v1.0: Full Type Listing}
\label{app:schema}

TRACE schema v1.0 defines two types. The root type:

\begin{quote}
\small
\begin{verbatim}
TraceRecord = {
  schema_version, writer_id, policy_version, claim_id,
  claim_text, claim_type, reasoning_family, argument_scheme,
  premises, warrants, role_states, evidence,
  source_credibility, counter_reasons, defeaters,
  context_frame, value_frame, embedded_subclaims,
  routing_decisions, criticisms, stage_results, attack_graph,
  audit_result, score_vector, failed_gates, missing, repair,
  cost, provenance, regret_score, revision_history,
  consumer_actions, final_status
}
\end{verbatim}
\end{quote}

For causal claims, the record specializes into the TRACE causal record:

\begin{quote}
\small
\begin{verbatim}
TraceCausalRecord = TraceRecord + {
  causal_rung, variables, graph_fragment, assumptions,
  missing_evidence, alternative_explanations, intervention,
  counterfactual_query, identifiability_status, bounds
}
\end{verbatim}
\end{quote}

\section{The Reference Writer: Stages, Algorithms, Cost, and Detail Tables}
\label{app:writer}

The reference writer constructs a \texttt{TraceRecord} from raw text in
eight stages, each holding a declared contract. The decomposition is a
reference architecture induced by the information dependencies among the
record's fields: a claim cannot be tested before it is typed, a scheme's
critical questions cannot be loaded before the scheme is identified, and an
attack graph cannot be adjudicated before its edges are audited. Whether
the staging earns its cost is settled by the ablations of
Section~\ref{sec:ocr-tracebench}. This appendix specifies the pipeline at
a glance, the two algorithms split at the record boundary, the modality
contract, and the cost model.

\paragraph{The pipeline at a glance.} Everything that follows elaborates
one eight-stage pipeline, shown in Figure~\ref{fig:ocr-pipeline} of the main text. Stages read
and write one shared artifact, the TRACE record; the two algorithms state
the executable contract, split at the record boundary.

The pipeline admits an implementation-level statement. TRACE is intended not
as a metaphor for evaluation, but as an executable methodology: each stage has
a declared input, output, state update, failure mode, and score contribution.
Given an input passage, the evaluator detects whether the passage contains a
reasoning relation, constructs a \texttt{TraceRecord}, decomposes the
passage into typed subclaims, routes each subclaim to the appropriate
reasoning-family standard, applies scheme-specific critical questions and
formal plug-ins where available, elicits and audits counter-reasons when the
escalation gate warrants, and emits a typed verdict with a score vector. This
turns the paper's survey of reasoning theories into an operational pipeline
for writing and auditing TRACE records; the sections that follow develop each line,
and Section~\ref{sec:ocr-endtoend} runs the whole algorithm on one sentence.
The statement is split at its natural seam: Algorithm~\ref{alg:ocr-trace}
turns raw text into a typed record, and Algorithm~\ref{alg:ocr-trace2} turns
the typed record into a verdict; the record is the only artifact that crosses
the boundary.

We call the executable form \textbf{TRACE}: \emph{Typed Reasoning Argument
Critique and Evaluation}. TRACE is not a truth oracle and not a replacement
for domain authority. Its output is a structured audit: what was claimed, what
kind of reasoning the claim requires, which evidence and warrants support it,
which defeaters remain unresolved, what verdict follows, and which error and
cost terms were incurred. The contribution of TRACE is to make reasoning
evaluation executable without pretending that all reasoning standards reduce
to one mechanism: symbolic checks, causal identification, argument-scheme
questions, model-mediated critique, retrieval, and human adjudication all
become plug-ins behind a shared record contract.

\begin{algorithm}[htbp]
\caption{TRACE, Part I: Detect, Intake, and Type (stages 0--2)}
\label{alg:ocr-trace}
\begin{algorithmic}[1]
\REQUIRE Input passage $x$, context $c$
\ENSURE Typed TRACE record $R$ with routed subclaims, or \textsc{No Argument}
\STATE $D \gets \mathcal{R}(x)$ \COMMENT{segment text; detect claim--support relations}
\IF{$D$ contains no support relation above threshold $\tau_R$}
    \RETURN \textsc{No Argument}
\ENDIF
\STATE Initialize $R$ with claim text, candidate premises, evidence links, context frame, value frame, and empty critique fields.
\STATE Apply CRIT--Toulmin intake to fill claim, grounds, warrants, backing, qualifiers, rebuttals, undercutters, counter-considerations, and burden fields.
\STATE Mark each role in $R$ as \textsc{Stated}, \textsc{Elicited}, \textsc{Reconstructed}, \textsc{Absent}, or \textsc{Contradicted}.
\STATE Decompose $R$ into typed subclaims $\{S_1,\ldots,S_k\}$.
\FORALL{subclaims $S_i$}
    \STATE Assign claim type, reasoning family, argument scheme, and confidence.
    \STATE Route $S_i$ to the corresponding family plug-in.
\ENDFOR
\RETURN $R$
\end{algorithmic}
\end{algorithm}

\begin{algorithm}[htbp]
\caption{TRACE, Part II: Question, Test, Elicit, Settle, and Score (stages 3--7)}
\label{alg:ocr-trace2}
\begin{algorithmic}[1]
\REQUIRE Typed record $R$ from Algorithm~\ref{alg:ocr-trace}, evidence ledger $\mathcal{E}$, stakes and budget
\ENSURE Completed record $R$ carrying verdict, score vector, failed gates, repair proposal, and cost
\FORALL{subclaims $S_i$ in $R$}
    \STATE Attach Walton-style critical questions and compute critical-question coverage.
    \STATE Route any critical question with a formal family counterpart to that plug-in; record the defect once, at the strongest applicable level.
    \STATE Apply the family-specific tests required by $S_i$: validity and soundness; sample and reference class; alternative explanations; Pearl rung, identification, and bounds; means--end fit; rule fit; or value/context fit.
    \IF{a hard flag is raised and declared stakes and remaining budget warrant escalation}
        \STATE Convene conditioned debate under the stopping rule to elicit counter-reasons, missing evidence, alternative explanations, undercutters, and rebuttals.
    \ENDIF
    \STATE Audit each proposed support or attack against the evidence ledger $\mathcal{E}$ and write pass, fail, bound, or defer status into $R$.
\ENDFOR
\STATE Build the attack graph $A_R$ from surviving supports, rebuttals, and undercutters.
\STATE Adjudicate $A_R$ with the applicable acceptability rule or external judge and write the verdict $V \in \{\textsc{Accept},\textsc{Qualify},\textsc{Revise},\textsc{Defer},\textsc{Reject}\}$ into $R$.
\STATE Write the score vector, failed gates, and cost into $R$: detection, type accuracy, scheme fit, evidence support, warrant explicitness, value transparency, alternative recall, overclaim rate, rung-collapse rate, bound violations, calibration gain, regret reduction, recurrence, and cost.
\STATE Write the missing and repair fields of $R$ from absent roles, failed critical questions, unresolved defeaters, missing evidence, and failed gates.
\RETURN $R$
\end{algorithmic}
\end{algorithm}

TRACE treats implementation modality as part of the contract. A plug-in may be
symbolic, statistical, model-mediated, retrieval-mediated, human-mediated, or
hybrid. The requirement is not that every plug-in be purely algorithmic; the
requirement is that every plug-in declare its input fields, output fields,
failure modes, confidence, cost, and audit trail. What makes the pipeline
automatic is the contract at each step: a detector must emit segments and
confidence; a router must emit claim types, families, and schemes; each plug-in
must emit pass, fail, bound, or defer status; and the adjudicator must write a
final status and metric vector back to the record. This is why the record, not
the transcript, is the central artifact.

\paragraph{Cost anatomy.} TRACE is affordable because it is a cascade: each
step is a filter, and the expensive machinery runs only on the fraction of
inputs that survives the filters ahead of it. Writing \(p_a\) for the
fraction of segments that are arguments and \(p_e\) for the fraction of
arguments whose flags, stakes, and budget trigger the debate escalation in
Algorithm~\ref{alg:ocr-trace2}, the expected cost per segment is
\[
\mathbb{E}[\text{cost}] \;\approx\; c_0 + p_a\,(c_{\mathrm{intake}} +
c_{\mathrm{tests}}) + p_a\, p_e\, c_{\mathrm{debate}},
\]
where the debate term dominates the constants. Detection is a small
classifier over every segment; intake, typing, and scheme selection are one
structured call per argument; and the gate keeps \(p_e\) small and
stakes-directed. Some family tests, such as proof checking, causal
identification, bounds, and graph acceptability, can be implemented
algorithmically. Others, such as practical-reasoning assessment,
source-credibility judgment, value-context fit, and alternative generation,
may be model-mediated, retrieval-mediated, human-mediated, or hybrid. Three
economies compound over episodes: stakes-proportional gating ties spend to
the expected regret of acting unevaluated; adjudicated claim patterns cache
through RLER; and Trivium compiles recurring error classes into cheap
formulation-gate flags and stage-3 questions.
TRACE therefore does not merely score an isolated argument; it accumulates
reusable reasoning-control policy. The cascade controls cost, not model
dependence: intake, typing, value-sensitive judgment, and debate remain
model-mediated or human-mediated where the contract requires them, and the
auditing caveats of the reconstruction algorithm apply to them unchanged.

The immediate design consequence is routing. A causal-sounding claim must be
routed to causal standards; a legal-sounding claim to legal and deontic
standards; a policy recommendation to practical reasoning, embedded empirical
claims, causal subclaims, and value warrants. A model that says \(X\) caused
\(Y\) has not yet made an evaluable causal claim unless the evaluator can tell
whether the claim is associational, interventional, or counterfactual. A model
that says a policy is justified has not yet made an evaluable practical
argument unless the evaluator can separate facts, causal effects, alternatives,
costs, institutional constraints, and value assumptions.

\paragraph{Detail tables.}
\label{sec:ocr-causal-instance}
\label{sec:ocr-operation}
The family plug-ins and the causal module mapping:

\begin{table}[htbp]
\centering
\small
\begin{tabular}{p{0.18\linewidth}p{0.34\linewidth}p{0.36\linewidth}}
\toprule
\textbf{Family plug-in} & \textbf{Primary tests} & \textbf{Typical outputs} \\
\midrule
Deductive & validity, soundness, formal or proof check & valid; invalid; unsound premise \\
\addlinespace[2pt]
Inductive/statistical & sample, reference class, calibration & supported pattern; biased sample; weak generalization \\
\addlinespace[2pt]
Abductive & explanatory coverage, alternatives, simplicity & best explanation; undercompared explanation \\
\addlinespace[2pt]
Causal & rung, identification, assumptions, bounds & identified; bounded; underidentified; defer \\
\addlinespace[2pt]
Practical & goals, means, alternatives, side effects, proportionality & justified action; revise; reject; defer \\
\addlinespace[2pt]
Legal/deontic & rule fit, authority, precedent, burden, value frame & meets standard; fails burden; jurisdiction dependent \\
\bottomrule
\end{tabular}
\caption{Family plug-ins in the typed reasoning pipeline. Every plug-in runs
through the same shared stages; causal reasoning is the worked instance, not
the whole architecture.}
\label{tab:ocr-plugins}
\end{table}

\begin{table}[H]
\centering
\small
\begin{tabular}{p{0.20\linewidth}p{0.36\linewidth}p{0.34\linewidth}}
\toprule
\textbf{Component} & \textbf{Role} & \textbf{Measurable Output} \\
\midrule
Pearl hierarchy & Type causal claims as association, intervention, or counterfactual & Rung accuracy; rung-collapse error rate \\
CRIT & Evaluate claim, reasons, evidence, credibility, rivals, and context & Invalid-support detection; overclaim reduction; source audit \\
SocraSynth & Elicit competing hypotheses through structured exchange & Hypothesis coverage; alternative-explanation recall \\
EVINCE & Measure evidence-seeking exchange and convergence or drift & Evidence gain; contradiction exposure; calibration shift \\
RCA & Diagnose why a causal judgment failed & Failure-localization accuracy; correction target \\
ERM & Score the epistemic cost of causal error & Regret score; severity-weighted error \\
RLER & Convert failures into revised reasoning policies & Repeated-error reduction; transfer after revision \\
Trivium & Evaluate improvement over time & Long-horizon regret reduction; temporal stability \\
\bottomrule
\end{tabular}
\caption{Causal reasoning as the worked plug-in inside the typed reasoning pipeline. CRIT is placed before debate because it defines the reasoning standard; SocraSynth and EVINCE are elicitation and measurement procedures.}
\label{tab:ocr-pipeline}
\end{table}

\paragraph{The enthymeme-reconstruction algorithm (stage 1).} Once the
roles are numbered, reconstruction stops being an art and becomes a
procedure. Number the roles $R_1$ claim, $R_2$ grounds, $R_3$ warrant,
$R_4$ backing, $R_5$ qualifier, $R_6$ rebuttal, and the extensions $R_7$
counter-considerations, $R_8$ undercutters, and $R_9$ presumption and
burden. For a given segment, each role is in exactly one of five states:
\emph{stated} (present in the text), \emph{elicited} (supplied by the
source in answer to a query), \emph{reconstructed} (supplied by the
evaluator and marked as such), \emph{absent} (marked empty after
elicitation and reconstruction both fail), or \emph{contradicted} (the
source's own statements are mutually inconsistent on this role). The procedure fills the roles in
dependency order, queries the source when one is available, and emits
the scores that the metric definitions of Section~\ref{sec:ocr-measurement}
already provide.

\begin{enumerate}[leftmargin=1.7em]
  \item \textbf{Claim.} Extract $R_1$. If no claim is found, the segment is
  not an argument; return it to detection.
  \item \textbf{Grounds.} Extract $R_2$. If absent, query the source:
  \emph{what supports this claim?} If elicitation fails, mark absent; an
  ungrounded claim is scored as assertion, not argument.
  \item \textbf{Warrant.} Test whether $R_2$ defeasibly licenses $R_1$
  without additions. If not, generate the minimal bridging proposition $W$
  by abduction and present it: \emph{your argument assumes $W$; is that
  your assumption?} Confirmed: elicited. Unconfirmed but plausible:
  reconstructed. No plausible $W$: hard gate; reject the step as a non
  sequitur.
  \item \textbf{Backing.} For each warrant, ask what supports it, and
  recurse: backing is itself an argument, and the recursive source audit
  applies to it.
  \item \textbf{Qualifier.} Read the stated modality; compute the strongest
  qualifier the grounds license; if the stated strength exceeds the
  licensed strength, raise the quantifier-mismatch flag, an overclaim
  signal.
  \item \textbf{Rebuttal.} Load the identified scheme's critical questions;
  every unanswered exception-type question becomes a candidate rebuttal
  condition, queried if a source is available.
  \item \textbf{Extensions.} Scan for concessive constructions ($R_7$),
  applicability challenges ($R_8$), and default or burden context ($R_9$);
  route $R_7$ and $R_8$ to the attack graph and $R_9$ to external
  adjudication.
  \item \textbf{Score.} Emit four reconstruction scores:
  $\mathrm{WarrantExplicitness}$, the support-weighted fraction of
  essential warrant roles that are stated or elicited;
  $\mathrm{CQCoverage}$, the fraction of applicable critical questions
  answered; quantifier consistency, a flag comparing the stated qualifier
  against the strongest the grounds license; and the charity delta
  $\Delta$, the support-weighted fraction of essential roles that had
  to be reconstructed. Weights reflect each role's support load. The reasonableness of the argument \emph{as
  stated} is reported as this vector under the gate-then-optimize rule;
  elicited roles improve the \emph{argument}, while reconstructed roles
  improve only the evaluator's \emph{reconstruction}, and $\Delta$ preserves
  the difference.
\end{enumerate}

Two remarks close the loop. First, the queries of steps 2, 3, and 6 are
CRIT's elenchus and dialectic in interactive form: the algorithm is the
Socratic method run against a schema, which is why its outputs land
directly in the TRACE record rather than in a side channel. Second,
steps 3 and 5 are themselves evaluator judgments; when the evaluator is a
language model, they inherit the reliability questions of any model
judgment, and sampled human audit of reconstructed warrants is the
corresponding control.


\section{The Extended Anatomy and the Contested Lists}
\label{app:anatomy}

\paragraph{The contested list.} No
completeness theorem exists at the level of natural argument, and the
literature contests the list in both directions: proposed additions include
the \emph{counter-consideration} of conductive
argument~\cite{wellman1971challenge,govier1987problems}, the
\emph{undercutting defeater} that attacks the warrant's applicability
rather than the claim~\cite{pollock1987defeasible}, and \emph{presumption
with burden of proof}~\cite{gordon2007carneades}, while Freeman would merge
the warrant into the grounds
altogether~\cite{freeman1991dialectics,hitchcock2006arguing}. A taxonomy
that serious scholars want to both extend and shrink is a useful schema, not
a closed enumeration.

Table~\ref{tab:ocr-extended-anatomy} assembles the resulting extended
anatomy: the six canonical roles, the three best-supported extensions,
their reading in the formal core, and the pipeline component that handles
each. Nothing on the extended list is homeless, and nothing requires the
six-role list to be closed: counter-considerations and undercutters are
resolved in the attack graph of the acceptability layer, and burden of
proof is a parameter of external adjudication.

\begin{table}[htbp]
\centering
\footnotesize
\begin{tabular}{p{0.17\linewidth}p{0.10\linewidth}p{0.30\linewidth}p{0.30\linewidth}}
\toprule
\textbf{Role} & \textbf{Status} & \textbf{Formal-core reading} & \textbf{Handled by} \\
\midrule
Claim & core & conclusion & record: \texttt{claim\_text}, \texttt{claim\_type} \\
\addlinespace[2pt]
Grounds & core & premises & record: \texttt{premises}, \texttt{evidence} \\
\addlinespace[2pt]
Warrant & core & inference rule & record: \texttt{warrants} \\
\addlinespace[2pt]
Backing & core & recursion: an argument whose conclusion supports the warrant & recursive backing/source audit \\
\addlinespace[2pt]
Qualifier & core & strength of the rule & quantifier and modality checks \\
\addlinespace[2pt]
Rebuttal & core & exception annotation on the rule & record: \texttt{defeaters} \\
\addlinespace[2pt]
Counter-consideration & extension & attacking argument, conceded and outweighed & attack graph (acceptability layer) \\
\addlinespace[2pt]
Undercutter & extension & attack on the rule's applicability & attack graph (attack typing) \\
\addlinespace[2pt]
Presumption, burden of proof & extension & default status and proof allocation & external adjudication; legal battery \\
\bottomrule
\end{tabular}
\caption{The extended anatomy under the two-tier view. Closure holds at the
formal tier, where one step has three components and therefore three attack
surfaces; the working tier is an open checklist whose known extensions are
all handled by existing pipeline components.}
\label{tab:ocr-extended-anatomy}
\end{table}

\section{Illustrations at Clinical and Policy Stakes}
\label{app:illustrations}
\label{sec:ocr-illustrations}

This section illustrates the record where mistakes are expensive: stylized
clinical records, a worked policy argument, and the typed evaluation that
replaces holistic scoring.

\subsection{Stylized Clinical TRACE Records}
\label{sec:ocr-clinical-examples}

The pipeline is easiest to understand in clinical reasoning, where causal mistakes have different consequences at different stages. Diagnosis begins with symptoms and uncertain causal hypotheses. Test selection asks which observation would reduce uncertainty. Treatment selection becomes an interventional and often counterfactual question. Follow-up asks whether the chosen intervention caused benefit, harm, tolerance, resistance, or recurrence. The following examples are stylized records used only to illustrate causal reasoning. They are not clinical advice, and they are not intended to reflect full clinical guideline pathways; they illustrate reasoning-record transitions.

\paragraph{Example 1: lung cancer treatment.}
Consider a stylized case in which a patient presents with persistent cough, weight loss, and blood-tinged sputum, and imaging reveals a lung mass. The initial claim is not yet ``this patient has lung cancer.'' It is a diagnostic causal claim: the observed symptoms and image may be produced by infection, inflammation, metastatic disease, or primary lung cancer. Pearl typing therefore places the early question near association and causal explanation, not yet treatment intervention. SocraSynth generates the differential diagnosis. CRIT rejects invalid jumps such as ``a lung mass implies cancer'' or ``symptom improvement after antibiotics proves infection.'' EVINCE then asks what evidence would distinguish the hypotheses: prior imaging, infection markers, biopsy, histology, stage, molecular profile, comorbidities, functional status, and patient preference.

Only after diagnosis and staging does the reasoning problem become primarily interventional: which treatment is expected to improve outcome under this patient's disease state and constraints? At this point the TRACE causal record changes from a diagnostic record to a treatment record. Candidate interventions may include surgery, radiation, systemic therapy, targeted therapy, immunotherapy, palliative symptom control, or clinical-trial referral. CRIT rejects treatment claims that ignore stage, histology, molecular evidence, toxicity, or the difference between tumor response and cure. EVINCE compares evidence for each intervention. ERM penalizes high-regret errors such as delayed biopsy, omitted molecular testing, unnecessary toxicity, or overconfident recommendation before sufficient staging evidence. If treatment fails or recurrence occurs, RCA localizes the failure: wrong diagnosis, incomplete staging, missing biomarker, toxicity underestimated, resistance not monitored, or goals of care mis-specified. RLER converts the failure into a revised policy, for example: do not recommend disease-control treatment before diagnosis, staging, and treatment-relevant evidence are explicit. Trivium then evaluates whether future cases show fewer missing-test, premature-treatment, or recurrence-monitoring errors.

The example shows why the pipeline must distinguish a valid causal claim from an actionable causal proposal; a second stylized record, early cognitive decline with treatment selection under asymmetric regret, follows the same transitions and is left to the record repository. A claim can be well supported and still fail to justify action if the intervention is infeasible, too risky, poorly aligned with patient goals, or insufficiently evidenced for the decision context. Operational causal reasoning therefore requires more than a correct explanation. It requires knowing which rung the system is on, what evidence is missing, when additional tests are justified, when treatment should be deferred, and how later outcomes should revise future reasoning.

\subsection{Evaluating Practical Arguments: A Worked Policy Example}
\label{sec:ocr-practical}

Consider the statement: \emph{ads can be deceiving, so ads should not appear in
children's programs}. This is not mainly a causal argument. It is a practical-policy
argument with embedded empirical, causal, and normative subclaims. A proper decomposition
is:

\begin{quote}
\small
\begin{tabular}{@{}p{0.05\linewidth}p{0.58\linewidth}p{0.27\linewidth}@{}}
P1. & Some ads can deceive. & \emph{empirical claim} \\
P2. & Children are vulnerable to deception. & \emph{empirical, developmental claim} \\
P3. & Exposure to ads in children's programs can harm children. & \emph{causal, interventional claim} \\
P4. & Removing ads from children's programs would reduce that harm. & \emph{causal, policy claim} \\
P5. & Protecting children from deception outweighs the costs of removing ads. & \emph{normative, value claim} \\
C. & Therefore, ads should not appear in children's programs. & \emph{normative policy conclusion} \\
\end{tabular}
\end{quote}

The primary scheme is the argument from negative consequences: if action $A$ permits
harmful consequence $H$, and $H$ should be avoided, then $A$ should be restricted or
replaced. But practical reasoning is not validated by empirical evidence alone; it also
requires a value premise. That is Hume's gap: facts do not by themselves yield
\emph{should}. The original argument is therefore incomplete, not necessarily false, and
the evaluation must say which. Seven dimensions make the evaluation systematic.

\paragraph{Claim-type classification.} The evaluator first identifies the conclusion type
(normative policy claim), the embedded subclaims (empirical, causal, value-based), and
the primary scheme (negative consequences, practical reasoning). This prevents applying
the wrong evaluator: a causal evaluator asks whether $X$ causes $Y$; a practical-policy
evaluator asks whether, given evidence, consequences, values, alternatives, and affected
stakeholders, the action is justified.

\paragraph{Premise support.} Each premise gets its own evidence question: are ads
sometimes deceptive (P1); are children less able than adults to recognize persuasive
intent (P2); does exposure cause measurable harm (P3); would removal reduce the harm
(P4); what costs and competing values would a ban impose (P5)? P2 is well supported:
industry self-regulation itself treats children as having limited knowledge, experience,
sophistication, and maturity, noting that younger children may not understand persuasive
intent or even that they are being advertised to~\cite{caru2021guidelines}.

\paragraph{Warrant explicitness.} The original sentence hides its warrant: \emph{because
children are vulnerable, society has a duty to protect them from deceptive persuasion.}
That is not a data claim; it is a normative bridge from evidence to policy, and an
argument that hides it cannot be evaluated until it is surfaced. Warrant explicitness is
low in the original and high after repair, and the score records the difference.

\paragraph{Quantifier and modality consistency.} The premise is weak and existential,
some ads \emph{can} deceive; the conclusion is strong and universal, all ads banned from
the context. Conclusion strength exceeding premise strength is the practical-argument
form of overclaim. The repaired argument needs stronger premises: children cannot
reliably distinguish persuasive intent; child-directed advertising predictably exploits
this vulnerability; screening only deceptive ads is insufficient or infeasible; therefore
broad restriction is justified.

\paragraph{Embedded causal typing.} Inside the practical argument sits an interventional
claim, P4: $do(\text{remove ads}) \rightarrow$ lower deception and harm. This subclaim is
extracted and routed to the causal plug-in of this paper: Pearl typing marks it
interventional; SocraSynth generates the alternatives (labeling, parental controls,
advertising literacy, platform rules); CRIT rejects the unsupported causal jump; EVINCE
gathers exposure and harm evidence; ERM penalizes overconfident policy when the causal
evidence is weak; RLER and Trivium track whether future policy arguments distinguish
exposure, deception, harm, and remedy.

\paragraph{Counter-consideration recall.} A practical argument is evaluated partly by
what it considered and rejected. Missing here: could ads be regulated rather than banned;
could educational and public-service advertising remain; would removing ads reduce access
to free programming; would children migrate to less regulated platforms; would parental
controls be more targeted? Pragma-dialectics makes this a general standard: argument
quality is judged not only by internal consistency but by whether the discussion
reasonably addresses the points at issue, the implicit premises, and the
scheme-appropriate criticisms~\cite{vaneemeren2004systematic}.

\paragraph{Value-context fit: DIKE and ERIS.} The value premise P5 is not resolvable by
evidence, and its evaluation is reasoning with declared criteria. DIKE asks which
fairness, protection, harm-prevention, rights, and regulatory norms apply: a declared,
versioned criteria set, the normative analog of $\mathcal{C}$. ERIS asks whose context is
missing: advertisers, parents, children of different ages, low-income families relying on
free content, educational broadcasters, different cultures and legal systems. The
argument is not fully evaluable without this context. In one jurisdiction a full ban may
be proportionate; in another, restricting manipulative formats, requiring disclosure, or
banning categories may fit better. An evaluation is complete only when it names the
criteria set and version it applied; the governance of those criteria is developed in the
volume's later chapters.

\subsection{From Holistic Scoring to Typed Argument Evaluation}
\label{sec:ocr-crit-typed}

The evaluation practice this implies is a typed extension of CRIT. The original CRIT question, does
this argument support the claim, remains necessary, but it is now refined by claim type, scheme, context, and value frame. The upgraded question is:

\begin{quote}
Given the claim type, argument scheme, context, evidence standard, and value frame, does
this argument provide sufficient support for this conclusion?
\end{quote}

Table~\ref{tab:ocr-crit-dimensions} gives the scoring dimensions that operationalize the
upgraded question. Each is checkable separately, which is what makes the evaluation
stronger than a single holistic score.

\begin{table}[H]
\centering
\small
\begin{tabular}{p{0.30\linewidth}p{0.62\linewidth}}
\toprule
\textbf{Dimension} & \textbf{Question it answers} \\
\midrule
Argument-type accuracy & Is the claim correctly classified as factual, causal, practical, normative, legal, analogical, or diagnostic? \\
\addlinespace[2pt]
Scheme fit & Is the argument using the right scheme: consequences, expert opinion, analogy, precedent, cause--effect, sign, rule, value, explanation? \\
\addlinespace[2pt]
Premise support & Are the explicit premises acceptable and evidenced? \\
\addlinespace[2pt]
Warrant explicitness & Are the hidden bridges between premise and conclusion surfaced? \\
\addlinespace[2pt]
Inference strength & Does the conclusion follow with the claimed strength? \\
\addlinespace[2pt]
Quantifier consistency & Does the conclusion overgeneralize beyond the premises? \\
\addlinespace[2pt]
Embedded-claim typing & Are embedded empirical, causal, and value claims separated and routed to their evaluators? \\
\addlinespace[2pt]
Counter-consideration recall & Are the major objections and alternatives considered? \\
\addlinespace[2pt]
Context fit & Does the argument adapt to audience, stakes, vulnerability, culture, law, and domain? \\
\addlinespace[2pt]
Value transparency & Are the value assumptions explicit rather than smuggled in? \\
\addlinespace[2pt]
Charity delta & How much reconstruction is required before the argument becomes strong? \\
\addlinespace[2pt]
Final status & Accept, accept with qualification, revise, defer, or reject. \\
\bottomrule
\end{tabular}
\caption{Typed argument evaluation: the upgraded CRIT scoring dimensions. Each dimension
is separately checkable; the final status extends the paper's three verdicts with the
graded outcomes practical arguments require.}
\label{tab:ocr-crit-dimensions}
\end{table}

Applied to the running example, the judgment as stated is: type, practical-policy
argument with embedded empirical and causal subclaims; scheme, argument from negative
consequences; quality, weak to moderate; reason, the conclusion may be defensible but the
argument is underdeveloped, with a hidden normative warrant, a quantifier jump, a missing
child-vulnerability premise, missing intervention evidence, and missing
counter-considerations. The repaired argument reads:

\begin{quote}
Because young children have limited ability to recognize persuasive intent, advertising
embedded in children's programs can exploit a vulnerable audience. If such exposure
predictably increases deception, unhealthy persuasion, or parent--child pressure, and if
less restrictive controls are insufficient, then restricting or banning ads in children's
programs is justified to protect children from unfair persuasion.
\end{quote}

The post-repair judgment: much stronger, because claim type, empirical support, causal
mechanism, value warrant, and policy scope are now explicit; still open, the age
threshold, the definition of an ad, the evidence of harm, the effectiveness of a ban,
exceptions for educational and public-service content, and migration to other platforms.
The distance between the two judgments is the charity delta, and it is itself a quality
signal about the source of the original argument.

The TRACE record of Section~\ref{sec:ocr-record} generalizes accordingly, with four
added fields: the scheme, the scheme's critical questions with their answers, the
warrants, and the embedded subclaims with their routing. The paper's summary statement
for this section is one sentence:

\begin{quote}
CRIT does not measure whether a sentence sounds reasonable; it types the claim,
identifies the argument scheme, reconstructs hidden warrants, evaluates embedded
subclaims under the right reasoning discipline, and scores whether the conclusion is
justified in context.
\end{quote}

\section{TRACE-Bench: Benchmark and Ablation Protocol}
\label{app:bench}
\label{sec:ocr-tracebench}

The operational metric definitions and their label sources:

\begin{table}[htbp]
\centering
\footnotesize
\begin{tabular}{p{0.16\linewidth}p{0.06\linewidth}p{0.17\linewidth}p{0.36\linewidth}p{0.06\linewidth}}
\toprule
\textbf{Metric} & \textbf{Unit} & \textbf{Label source} & \textbf{Formula or scoring rule} & \textbf{Gate} \\
\midrule
DetectionF1 & $[0,1]$ & annotated reasoning and non-reasoning segments & segmentation F1 combined with reasoning-segment classification F1 & pre \\
\addlinespace[2pt]
TypeAcc & $[0,1]$ & expert claim-type labels & macro-F1 of predicted against labeled claim types & op. \\
\addlinespace[2pt]
SchemeFit & $[0,1]$ & annotated scheme labels & $\mathrm{SchemeF1} \times \mathrm{CQCoverage}$, where CQCoverage is answered over applicable critical questions & no \\
\addlinespace[2pt]
EvidenceSupport & $[0,1]$ & evidence-link annotations or expert ordinal score & precision and recall of premise-to-evidence links, or normalized expert ordinal score & no \\
\addlinespace[2pt]
SourceTrace\-Completeness & $[0,1]$ & source audit & traced sources over cited or asserted external claims & no \\
\addlinespace[2pt]
SourceReliability\-Score & $[0,1]$ & credibility rubric & normalized expert or rubric credibility score of traced sources & no \\
\addlinespace[2pt]
Warrant\-Explicitness & $[0,1]$ & warrant annotations & hidden warrants surfaced over essential warrants & no \\
\addlinespace[2pt]
Value\-Transparency & $[0,1]$ & value-premise audit & declared value premises over required value premises & no \\
\addlinespace[2pt]
AltRecall & $[0,1]$ & expert or oracle alternative set & recall of surfaced alternatives against the reference set & no \\
\addlinespace[2pt]
CalibrationGain & signed & outcome labels & $\mathrm{Brier}_{\mathrm{pre}} - \mathrm{Brier}_{\mathrm{post}}$ (or the same difference in expected calibration error); signed, normalized before aggregation as $\mathrm{clip}((\mathrm{gain}+1)/2, 0, 1)$ & no \\
\addlinespace[2pt]
UnsupportedRate & $[0,1]$ & evidence audit & accepted premises without audited evidence, over accepted premises & yes \\
\addlinespace[2pt]
OverclaimRate & $[0,1]$ & modality and rung audit & conclusions whose quantifier, modality, or causal rung exceeds premise support, over all conclusions & yes \\
\addlinespace[2pt]
RungCollapseRate & $[0,1]$ & rung labels & causal claims concluding at rung 2 or 3 from rung-1 evidence without identification support, over causal claims & yes \\
\addlinespace[2pt]
BoundViolation\-Rate & $[0,1]$ & identification analysis & rung-2/3 verdicts tighter than the identifiable bounds, over rung-2/3 verdicts & yes \\
\addlinespace[2pt]
RegretReduction & cost units & ERM cost matrix & expected error cost before evaluation minus expected error cost after evaluation & no \\
\addlinespace[2pt]
RepeatedErrorRate & $[0,1]$ & RCA error classes & episodes repeating a previously diagnosed error class, over later episodes & no \\
\addlinespace[2pt]
CostNorm & $[0,1]$ & execution logs and budget & $\min(1, \mathrm{observed\ cost}/\mathrm{budgeted\ cost})$, where observed cost aggregates tokens, turns, wall-clock time, retrieval and model calls, and human-review burden & budget \\
\bottomrule
\end{tabular}
\caption{Operational definitions for the writer and record-quality dimensions. Gated
metrics participate in the hard gates of the final status; the rest are
optimized on the residual and reported in the vector. Provenance
(SourceTraceCompleteness) is objective and separated from credibility
(SourceReliabilityScore), which is partly subjective.}
\label{tab:ocr-metric-defs}
\end{table}

An implementation-level methodology needs an implementation-level benchmark.
TRACE-Bench is the benchmark protocol implied by the pipeline: it does not ask
only whether the final answer is correct. It asks whether the evaluator
detected the argument, reconstructed its roles, typed the claim, routed the
subclaims to the right standards, asked the right critical questions, invoked
the right plug-ins, surfaced the right defeaters, and assigned the right
verdict and repair. The benchmark is designed to evaluate both the full TRACE
implementation and each stage-level contract separately.

Each TRACE-Bench item is a short passage, usually one to five sentences,
paired with a context and optional evidence ledger. Items include non-arguments,
single-family arguments, and mixed arguments. The gold record contains: the
reasoning segment; claim and subclaims; premises, warrants, backing,
qualifier, rebuttal, counter-considerations, undercutters, and burden fields;
claim type; reasoning family; argument scheme; applicable critical questions;
family-specific checks; evidence status; unresolved defeaters; final status;
repair; and score vector. The same item can therefore evaluate both local
operators and end-to-end judgment.

\begin{table}[htbp]
\centering
\footnotesize
\begin{tabular}{p{0.16\linewidth}p{0.31\linewidth}p{0.41\linewidth}}
\toprule
\textbf{Task} & \textbf{Gold output} & \textbf{Primary metrics} \\
\midrule
T0 Detection & reasoning vs. non-reasoning segments; segment boundaries & DetectionF1; false-argument rate; missed-argument rate \\
\addlinespace[2pt]
T1 Anatomy & claim, grounds, warrants, qualifier, rebuttal, undercutters, burden & role F1; WarrantExplicitness; charity delta \\
\addlinespace[2pt]
T2 Typing and routing & claim type, reasoning family, scheme, mixed-subclaim decomposition & TypeAcc; family macro-F1; routing accuracy \\
\addlinespace[2pt]
T3 Scheme questions & applicable Walton critical questions and answers & SchemeFit; CQCoverage; double-count prevention \\
\addlinespace[2pt]
T4 Family plug-ins & validity, sample/reference class, alternatives, Pearl rung, identification, means--end fit, rule/value fit & plug-in accuracy; RungCollapseRate; BoundViolationRate; OverclaimRate \\
\addlinespace[2pt]
T5 Defeaters and alternatives & counter-reasons, missing evidence, undercutters, rebuttals, attack graph & AltRecall; defeater recall; attack-edge F1 \\
\addlinespace[2pt]
T6 Verdict and repair & accept, qualify, revise, defer, reject, or no argument; repaired claim & VerdictAcc; WrongAcceptRate; DeferQuality; RepairUsefulness \\
\addlinespace[2pt]
T7 Cost and learning signal & token, turn, retrieval, tool, and human-review cost; repeated-error labels & CostNorm; RegretReduction; RepeatedErrorRate \\
\bottomrule
\end{tabular}
\caption{TRACE-Bench evaluates the whole typed-reasoning methodology and its
operators. Unlike answer-only benchmarks, it scores the intermediate record:
detection, anatomy, routing, scheme questions, family tests, defeaters,
verdict, repair, and cost.}
\label{tab:ocr-tracebench-tasks}
\end{table}

The benchmark should be deliberately mixed. Deductive items test validity and
soundness; inductive items test sample and reference-class errors; abductive
items test alternative-explanation recall; causal items test association,
intervention, counterfactuals, confounding, identification, and bounds;
practical-policy items test means--end fit, side effects, alternatives, and
value warrants; legal or deontic items test rule fit, authority, precedent,
burden, and institutional context. Non-argument passages are necessary because
a detector that treats every fluent statement as an argument will look strong
on downstream tasks while failing at intake.

A minimal useful TRACE-Bench can be built in three tiers. Tier 1 uses synthetic
and templated examples whose gold decomposition is controlled. Tier 2 adapts
existing annotated argument-mining and causal-reasoning datasets into the
TraceRecord schema. Tier 3 uses expert-written mixed arguments in medicine,
policy, law, science, and everyday explanation, because the central challenge
is mixed reasoning: a single paragraph may contain empirical, causal,
practical, and value claims at once. Annotation should be redundant: at least
two annotators fill the record independently, disagreements are adjudicated,
and the adjudicated record becomes the gold ledger.

The ablation suite tests whether the architecture matters. The full system is
TRACE. Baselines include direct answer, chain-of-thought only, self-consistency,
LLM-as-judge holistic scoring, single-pass record emission in which one
structured call writes the entire TraceRecord, debate without typed
routing, CRIT without family plug-ins, and family plug-ins without Walton
critical questions. Two baselines carry special weight. Single-pass record
emission is the null hypothesis for the staged writer: it conforms to the
schema, so if it matches TRACE on record quality and error attribution at
comparable cost, the decomposition is unjustified
(Section~\ref{sec:ocr-falsify}). Debate without typed routing, scored
holistically, is the volume's own earlier evaluation mechanism, so this row
is the volume auditing its predecessor with its own methods. The important
ablations remove one control surface at a time.

\begin{table}[htbp]
\centering
\footnotesize
\begin{tabular}{p{0.25\linewidth}p{0.58\linewidth}}
\toprule
\textbf{Ablation} & \textbf{Expected failure mode} \\
\midrule
No detector & Non-arguments are forced into argument schemas; false positives contaminate all downstream scores. \\
\addlinespace[2pt]
No anatomy & Hidden warrants, qualifiers, rebuttals, and burdens disappear; verdicts become fluent but unauditable. \\
\addlinespace[2pt]
No family routing & Deductive, causal, practical, and legal claims receive the wrong standards; type errors propagate. \\
\addlinespace[2pt]
No schemes or critical questions & Critique becomes improvised; defeater recall and counter-consideration coverage fall. \\
\addlinespace[2pt]
No causal plug-in & Correlation-to-cause failures are noticed rhetorically but not adjudicated by rung, identification, or bounds. \\
\addlinespace[2pt]
No evidence ledger & Convergence and confidence cannot be separated from rhetoric, sycophancy, or agreement pressure. \\
\addlinespace[2pt]
No attack graph or external adjudication & Debate consensus is mistaken for acceptability; defeated claims can survive by majority or fluency. \\
\addlinespace[2pt]
No gates; scalar only & A high aggregate score can hide unsupported premises, rung collapse, bound violations, or overclaims. \\
\addlinespace[2pt]
No cost term & Overlong reasoning and redundant debate appear beneficial even when evidence gain has saturated. \\
\bottomrule
\end{tabular}
\caption{Ablation plan for TRACE-Bench. Each ablation removes one design
commitment of the paper and predicts the failure mode it should expose. A
useful benchmark must show not only that TRACE improves final verdicts, but
also which stage is responsible for the improvement.}
\label{tab:ocr-tracebench-ablations}
\end{table}

The primary end-to-end measures are \emph{WrongAcceptRate} and
\emph{DeferQuality}. WrongAcceptRate is the fraction of unsupported,
overclaiming, or formally invalid arguments that receive \textsc{Accept} or
\textsc{Qualify}. DeferQuality asks whether a defer verdict names the missing
evidence or assumption that would actually tighten, flip, or justify the
claim when supplied. These two measures capture the practical value of the
pipeline: it should not merely sound critical; it should prevent wrong
acceptance and make uncertainty actionable.

A useful reporting scalar is record utility:

\begin{equation}
\begin{aligned}
\mathrm{RecordUtility} ={}&
\alpha_1 \mathrm{TypeAcc}
+ \alpha_2 \mathrm{SchemeFit}
+ \alpha_3 \mathrm{EvidenceSupport}
+ \alpha_4 \mathrm{AltRecall} \\
&+ \alpha_5 \mathrm{CalibrationGain}
+ \alpha_6 \mathrm{WarrantExplicitness} \\
&+ \alpha_7 \mathrm{ValueTransparency}
+ \alpha_8 \mathrm{RegretReduction} \\
&- \beta_1 \mathrm{UnsupportedRate}
- \beta_2 \mathrm{OverclaimRate} \\
&- \beta_3 \mathrm{RungCollapseRate}
- \beta_4 \mathrm{RepeatedErrorRate} \\
&- \beta_5 \mathrm{CostNorm}.
\end{aligned}
\label{eq:ocr-reasoning-quality}
\end{equation}

The weights should be task dependent. In low-risk explanatory tasks, evidence gain and alternative recall may dominate. In high-stakes action settings, overclaim reduction, regret reduction, and recurrence control should receive higher weight.

The Cost term is not incidental. Reasoning models routinely produce redundant and overlong traces, and surveys of efficient reasoning document substantial token expenditure for little accuracy gain~\cite{chen2025overthinking,sui2025stopoverthinking}. Cost-aware quality metrics and the stopping rules of Section~\ref{sec:ocr-stopping} are the evaluator's countermeasures to overthinking.

\paragraph{Gate, then optimize.} Equation~\eqref{eq:ocr-reasoning-quality} should be read as a reporting device, not a training target. Optimized directly, a scalarized score invites Goodhart behavior: alternative recall can be inflated with junk alternatives to offset overclaim penalties, and cost terms can be gamed by early stopping. The operational form is lexicographic: hard gates on the non-negotiables (rung discipline, the unsupported-premise rate, and bound violations, per Proposition~\ref{prop:ocr-rungbound}), followed by scalar optimization on the residual. As a reporting rule, the scalar is never reported without the vector and the list of failed gates: the scalar is a dashboard summary, not the scientific result. The weights \(\alpha_i, \beta_j\) are governance decisions: versioned, audited, and owned by the governance layer (DIKE and ERIS) of the companion volume. All component terms should be normalized to \([0,1]\) before aggregation.

\section{Convergence Validity: Metrics, Imported Propositions, and an Open Problem}
\label{app:convergence}

The six validity metrics below make convergence validity measurable; the
companion volume computes them. The propositions that follow are the formal
spine: imported classical results stated as theorems, their transfer to
multi-LLM systems stated as design principles, and the open problem the
experiments are designed to inform.

\subsection{The Validity Metrics}
\label{sec:ocr-validity-metrics}

Let an episode consist of turns \(t=1,\dots,T\); let \(b_t\) denote the
consensus-relevant belief state after turn \(t\) (for a binary claim, a
probability); let \(E_t \subseteq \mathcal{E}\) be the evidence units newly
admitted to the ledger at turn \(t\) after CRIT audit at threshold
\(\theta\)~\cite{chang2023crit}; and let \(V\) be the final verdict.

\begin{enumerate}[leftmargin=1.7em]
  \item[\textbf{M1}] \textbf{Procedural invariance (convergence
  robustness).} Re-run the episode under a perturbation set \(\Pi\) that
  ought not to matter: role reassignment, speaking order, persona relabeling,
  sampling seed. Define
  \(\mathrm{CR} = 1 - \Pr_{\pi \in \Pi}[\,V^{\pi} \neq V\,]\).
  A consensus that flips with seating order is rhetorical, not epistemic.
  \item[\textbf{M2}] \textbf{Warranted convergence ratio.} Decompose belief
  movement by turn and call a movement \emph{warranted} when it co-occurs
  with newly admitted evidence:
  \[
  \mathrm{WCR} \;=\;
  \frac{\sum_{t}\,\lvert b_t - b_{t-1}\rvert\;
        \mathbf{1}[\,E_t \neq \varnothing\,]}
       {\sum_{t}\,\lvert b_t - b_{t-1}\rvert}.
  \]
  The complement \(1-\mathrm{WCR}\) is a per-episode \emph{sycophancy
  index}: concession without evidence, the failure mode documented
  empirically in multi-agent
  debate~\cite{wynn2025talk,yao2025peacemaker}.
  \item[\textbf{M3}] \textbf{Evidence-counterfactual sensitivity.} Ablate
  ledger items and re-adjudicate: the verdict should move when decisive
  evidence is removed and stay fixed when rhetoric-only turns are removed or
  paraphrased. This is the debate analogue of chain-of-thought faithfulness
  tests by corruption and truncation~\cite{lanham2023measuring}. A verdict
  insensitive to its own evidence is fluent consensus, now detectable.
  \item[\textbf{M4}] \textbf{Dual-phase divergence signature.} Let \(D_t\)
  be a cross-agent divergence (Jensen--Shannon or Wasserstein) over
  predictive distributions, as tracked by EVINCE~\cite{chang2024evince}. A
  healthy episode shows an \emph{exploration} phase in which \(D_t\) rises or
  holds while the hypothesis set grows, then a \emph{contraction} phase in
  which the fall of \(D_t\) correlates with evidence arrivals \(E_t\).
  Monotone contraction from the first turn with a static hypothesis set is
  premature collapse: degeneration of thought~\cite{liang2024divergent} as a
  measurable waveform rather than an anecdote.
  \item[\textbf{M5}] \textbf{Effective panel size.} With mean pairwise error
  correlation \(\bar\rho\) among \(n\) agents, estimated on a calibration set
  of labeled claims, define
  \(n_{\mathrm{eff}} = n / \bigl(1 + (n-1)\bar\rho\bigr)\).
  Jury-style aggregation benefits require \(n_{\mathrm{eff}} \gg
  1\)~\cite{ladha1992condorcet}; same-model panels have \(\bar\rho \to 1\)
  and hence \(n_{\mathrm{eff}} \to 1\), a one-line account of why
  homogeneous debate underperforms and heterogeneous panels are the
  remedy~\cite{zhang2025stop}. Report \(n_{\mathrm{eff}}\), not \(n\).
  \item[\textbf{M6}] \textbf{External-judge gap.} Give the ledger, and only the
  ledger, to an adjudicator that did not participate: a different-family
  model, a formal checker, or CRIT run cold. Measure the gap between its
  verdict and the participants' consensus. A real convergence survives being
  re-derived from its own evidence by a stranger.
\end{enumerate}

The empirical record of reasoning-as-debate methods, ordered by the four
debate-ness conditions of the main text:

\begin{table}[htbp]
\centering
\footnotesize
\begin{tabular}{p{0.17\linewidth}p{0.07\linewidth}p{0.09\linewidth}p{0.10\linewidth}p{0.11\linewidth}p{0.30\linewidth}}
\toprule
\textbf{Method} & \textbf{D1} & \textbf{D2} & \textbf{D3} & \textbf{D4} &
\textbf{Characteristic behavior} \\
\midrule
Chain-of-thought & -- & -- & -- & -- & monologue; gains are computational
decomposition, not adversarial testing; chains may be post
hoc~\cite{turpin2023language,lanham2023measuring} \\
\addlinespace[2pt]
Self-consistency & -- & -- & $\sim$ & vote & ensemble, not debate; variance
reduction with fully correlated priors~\cite{wang2023selfconsistency} \\
\addlinespace[2pt]
Self-critique & nominal & -- & -- & -- & debate with a mirror; self-correction
fails without external feedback~\cite{huang2024self} \\
\addlinespace[2pt]
Same-model multi-agent debate & \checkmark & $\sim$ & -- & -- &
gains on factuality~\cite{du2023multiagent} but degeneration of thought,
conformity, and consensus without
correction~\cite{liang2024divergent,weng2025conformity,choi2025debatevote} \\
\addlinespace[2pt]
SocraSynth / EVINCE & \checkmark & \checkmark & engineered & CRIT +
information dials & contentiousness restores D2; conditional statistics
engineer D3; CRIT and divergence monitors supply
D4~\cite{chang2024socrasynth,chang2024evince,chang2024pathagi1} \\
\addlinespace[2pt]
Prover--verifier debate & \checkmark & \checkmark & \checkmark & judge &
complexity-theoretic verification guarantees~\cite{irving2018debate,browncohen2023doubly,browncohen2025prover} \\
\bottomrule
\end{tabular}
\caption{The debate-ness profile D1--D4 orders reasoning-as-debate methods
and explains the mixed empirical record: exchange helps where D2--D4 are
enforced and disappoints where they are simulated.}
\label{tab:ocr-debateness}
\end{table}

\subsection{Imported Propositions and the Open Problem}

\begin{proposition}[Aumann agreement]
\label{prop:ocr-aumann}
If agents share a common prior and truthfully exchange posteriors that
become common knowledge, they cannot agree to disagree, and iterated
communication converges in finitely many rounds regardless of the truth of
the claim \cite{aumann1976agreeing,geanakoplos1982disagree}.
\end{proposition}

\paragraph{Design principle: consensus is weak evidence under correlated
generators.} The transfer of Proposition~\ref{prop:ocr-aumann} to multi-LLM
debate is a modeling assumption, not a theorem. Agents instantiated from a
common base model are strongly correlated belief generators that only
approximate the common-prior setting. Under that assumption, the likelihood
ratio of the event ``the debate converged'' with respect to the truth of the
claim should be treated as close to uninformative unless the convergence is
accompanied by ledger movement, external adjudication, and procedural
invariance: consensus alone is not evidence. Evidential weight resides in
the ledger and the belief trajectory, not in the agreement.

\begin{proposition}[Diversity, not head-count]
\label{prop:ocr-diversity}
Under the standard equal-variance, equal-correlation approximation,
majority aggregation over \(n\) agents with mean pairwise error correlation
\(\bar\rho\) behaves as an independent panel of size
\(n_{\mathrm{eff}} = n/(1+(n-1)\bar\rho)\), and jury-theorem guarantees
weaken accordingly \cite{ladha1992condorcet}. The reduction to a single
mean correlation is a design approximation, not an exact law of arbitrary
error dependence, but the direction of the effect is general. Corollary: a
homogeneous multi-agent debate approximates a single agent, and the marginal
value of an added agent is the marginal decorrelation it contributes.
\end{proposition}

\begin{proposition}[Verification asymmetry]
\label{prop:ocr-verification}
In the prover--verifier formalization, debate with optimal play permits a
polynomial-time judge to decide any problem in PSPACE
\cite{irving2018debate}; doubly-efficient debate makes the honest strategy
itself efficient \cite{browncohen2023doubly}, and prover--estimator debate
addresses obfuscated arguments \cite{browncohen2025prover}. Debate is
therefore not a brainstorming heuristic but a verification protocol: it
extends the reach of a bounded judge beyond anything solitary deliberation at
the judge's capacity can attain. These guarantees apply to formal debate
protocols with optimal play and specified judge models, not to ordinary
multi-agent LLM discussion.
\end{proposition}

\begin{openproblem}[Convergence to the ideal aggregator under engineered
diversity]
\label{op:ocr-engineered}
Characterize the conditions on role conditioning (contentiousness schedules,
persona assignment, corpus separation) under which the consensus of a
conditioned multi-LLM debate converges to the ideal aggregator
\(b^{*}(\mathcal{E})\); that is, when does \emph{simulated} prior diversity
purchase the aggregation benefits of \emph{real} prior diversity? A stylized
sufficient condition would require the conditioned agents' likelihood
contributions to be conditionally independent given the claim; whether and
when role conditioning achieves this is an empirical question that the
validity metrics of Section~\ref{sec:ocr-validity-metrics} are designed to
test.
\end{openproblem}

\section{Usage Scenario: Mnemosyne Memory Admission}
\label{app:mnemosyne}

Mnemosyne enforces the TRACE discipline at the memory boundary: nothing
enters durable memory without a TRACE record carrying a verdict. The gate
is one function, \texttt{admit(record)}, mapping the record's
\texttt{final\_status} and \texttt{failed\_gates} to one of four
outcomes: commit, commit with qualifier, quarantine, or reject. Quarantined
records persist with their verdicts, retrievable but marked, so a later
consumer can distinguish an unproven claim from an absent one.

The music-lessons record of the worked trace
(Section~\ref{sec:ocr-endtoend}) walks the gate as follows. Subclaim
\(S_1\), the observed association, arrives with status \textsc{Accept}
under an inductive qualifier and commits with that qualifier attached.
Subclaim \(S_2\), the causal claim, arrives as \textsc{Defer} with
\texttt{missing} naming randomized or quasi-experimental assignment; it
is quarantined, and the named missing evidence is registered as a trigger.
Subclaim \(S_3\), the policy conclusion, arrives as \textsc{Reject}
with its repair proposal; the rejection and the repair are logged, not
discarded, because Trivium's regret accounting consumes them.

Two behaviors then depend on fields no scalar carries. Defer is
\emph{prospective memory}: when evidence matching the trigger arrives, a
lottery-based school assignment study, say, the quarantined record is
re-adjudicated; if the verdict flips to \textsc{Accept}, the record
commits and the trigger retires, and if it hardens to \textsc{Reject},
the quarantine converts, in either case exercising the actionable sense of
the explanation claim (Section~\ref{sec:ocr-falsify}). Reconsolidation
is \emph{re-audit}: if a committed record's premise is later defeated,
belief revision propagates through \texttt{revision\_history} rather
than silently overwriting, and every downstream record whose
\texttt{provenance} cites the defeated premise is queued for re-audit.
The measured quantities are admission precision and recall against expert
judgment, stale-memory correction rate, and defer quality, the fraction of
defers whose named evidence, when supplied, in fact changes the verdict.

\section{Usage Scenario: MACI Plan Gating}
\label{app:maci}

MACI enforces the TRACE discipline at the commitment boundary of a
multi-agent workflow, under the division of labor stated in
Section~\ref{sec:ocr-falsify}: structural admission, whether the workflow is
well formed, its dependencies acyclic, and its resource and temporal
constraints satisfiable, belongs to the plan validators. TRACE gates the
justificatory residue, the claims a plan rests on that constraint checking
cannot reach.

The flood-rescue vignette of Section~\ref{sec:ocr-flood-vignette} makes the
division concrete. The structural validator admits all three candidates: a
northern boat dispatch, a bounded drone verification, and a southern boat
dispatch. The boat exists, its capacity is sufficient, the destination and
deadline are declared, and each route is syntactically usable by the planner.
The validator therefore does not reject the northern branch. TRACE reads the
record licensing that branch and finds a different defect: the reported
success value is high, but model support is below policy and the OOD score is
above policy. MACI writes \textsc{Hold} to \texttt{consumer\_actions}, cites
$R_N^{(1)}$, and leaves the alternative branches available.

The \texttt{repair} field of $R_N^{(1)}$ names a current route observation.
The planner proposes the survey-drone action, and a separate practical record
$R_V^{(1)}$ clears or qualifies that bounded information-gathering step. When
the drone reports the debris obstruction, the repository appends
$R_N^{(2)}$ rather than editing the earlier forecast. The
\texttt{revision\_history} link identifies which commitments depended on the
failed route premise, so MACI invalidates the northern branch only. The South
Detour remains structurally valid; after its own predictive and practical
record passes policy and the Incident Commander supplies external approval,
MACI writes \textsc{Clear} for the southern dispatch with the exact
authorizing record version.

The measured quantities are the prevented-error rate on plans that pass
structural validation yet fail justification, false-hold rate, added latency,
repair success, and branch-localization accuracy. The claim defended is still
the explanation and control-interface claim, not a rescue-performance claim:
every held, qualified, or cleared commitment arrives with why, on what
evidence, under which policy, and what later observation changed it. A
controlled gated-versus-ungated implementation is future work.

\section*{Acknowledgments}
The photorealistic flood search-and-rescue scene
(Figure~\ref{fig:ocr-flood-scene}) was generated with OpenAI's GPT-4o image
generation. All schematic and lifecycle diagrams are the authors' own vector
figures. The generated image is an illustration only; it depicts no real
person, place, or deployment.

\bibliographystyle{plainnat}
\bibliography{Trace_Schema}
\end{document}